\documentclass[preprint,authoryear,3p]{elsarticle}

\usepackage{enumitem}
\usepackage[most]{tcolorbox}
\usepackage{multirow}
\usepackage{pdflscape}
\usepackage{amssymb,amsthm,amsmath}
\usepackage{caption}
\usepackage{subcaption}
\usepackage{comment}
\usepackage{url}
\usepackage[english]{babel}
\usepackage[utf8]{inputenc}
\usepackage{algorithm}
\usepackage{algorithmic}
\usepackage{makecell}
\usepackage[dvipsnames]{xcolor}
\graphicspath{{figures/}} 
\usepackage{nomencl}
\usepackage{booktabs}
\makenomenclature

\usepackage{etoolbox}
\renewcommand\nomgroup[1]{%
  \item[\bfseries
  \ifstrequal{#1}{P}{Vehicle parameters}{%
  \ifstrequal{#1}{L}{Vehicle lateral control}{%
  \ifstrequal{#1}{R}{Reinforcement learning algorithm}{%
  \ifstrequal{#1}{A}{Actions and reward function}{%
  \ifstrequal{#1}{M}{Model predictive control}{%
  \ifstrequal{#1}{H}{Path planning}{}}}}}}%
]}

\makeatletter
\renewcommand\paragraph{\@startsection{paragraph}{4}{\z@}%
            {-2.5ex \@plus -1ex \@minus -.25ex}%
            {1.25ex \@plus .25ex}%
            {\normalfont\normalsize\itshape}}
\makeatother

\usepackage{color}

\journal{ }

\begin{document}

\begin{frontmatter}



\title{Federated Hierarchical Reinforcement Learning for Adaptive Traffic Signal Control\\
}


\author[cu]{Yongjie Fu}
\author[um]{Lingyun Zhong}
\author[ee]{Zifan Li}
\author[cu,dsi]{Xuan Di\corref{cor}}
\ead{sharon.di@columbia.edu}

\cortext[cor]{Corresponding author. Tel.: +1 212 853 0435;}
\address[cu]{Department of Civil Engineering and Engineering Mechanics, Columbia University}
\address[um]{Department of Civil and Environmental Engineering, University of Michigan}
\address[dsi]{Center for Smart Cities, Data Science Institute, Columbia University}
\address[ee]{Department of Electrical Engineering, Columbia University}
            
\begin{abstract}

Multi-agent reinforcement learning (MARL) has shown promise for adaptive traffic signal control (ATSC), enabling multiple intersections to coordinate signal timings in real time. However, in large-scale settings, MARL faces constraints due to extensive data sharing and communication requirements. Federated learning (FL) mitigates these challenges by training shared models without directly exchanging raw data, yet traditional FL methods such as FedAvg struggle with highly heterogeneous intersections. Different intersections exhibit varying traffic patterns, demands, and road structures, so performing FedAvg across all agents is inefficient. To address this gap, we propose Hierarchical Federated Reinforcement Learning (HFRL) for ATSC. HFRL employs clustering-based or optimization-based techniques to dynamically group intersections and perform FedAvg independently within groups of intersections with similar characteristics, enabling more effective coordination and scalability than standard FedAvg.Our experiments on synthetic and real-world traffic networks demonstrate that HFRL consistently outperforms decentralized and standard federated RL approaches, and achieves competitive or superior performance compared to centralized RL as network scale and heterogeneity increase, particularly in real-world settings. The method also identifies suitable grouping patterns based on network structure or traffic demand, resulting in a more robust framework for distributed, heterogeneous systems.

\end{abstract}

\begin{keyword}
Federated Learning \sep Reinforcement learning \sep Adaptive Traffic Signal Control \sep Hierarchical Framework.
\end{keyword}

\end{frontmatter}


\section{INTRODUCTION}



Traffic congestion has become a pressing issue in urban areas, exacerbated by the continuous growth in the number of vehicles. In 2023, drivers in New York City lost an average of $101$ hours due to traffic congestion, causing approximately \$9.1 billion in lost productivity \citep{NYPostTraffic2024}. Adaptive Traffic Signal Control (ATSC) systems have emerged as a promising solution to improve the efficiency of traffic flow by coordinating signal operations across multiple intersections in real time. By dynamically adjusting the traffic signal phase, ATSC aims to minimize delays and congestion, providing a more responsive traffic management system. Optimization methods are widely used in ATSC \citep{mohebifard2019cooperative, mohebifard2019optimal, yin2023real, zheng2024integrated}.

Recent years have seen a growing amount of literature that applies learning methods for ATSC, with the aim of learning the stochastic and unknown dynamics of traffic environments. 
Reinforcement learning (RL), in particular, has been widely applied to ATSC \citep{li2024cooperative, MO2022103728, zhang2024survey, aslani2017adaptive, wang2024large}. 
Traffic lights are modeled as intelligent agents that cooperate or optimize predefined traffic metrics in a decentralized manner while learning traffic dynamics. Specifically, RL-based approaches can dynamically adapt to real-time traffic conditions, making them well suited to handle variations in flow rates and unexpected congestion patterns. 
Moreover, these methods often outperform traditional rule-based solutions by continuously learning and refining control strategies based on ongoing feedback from the traffic environment \citep{wei2019presslight, wang2024large}. 

However, applying RL to ATSC presents several challenges. One potential concern arises when RL models rely on fine-grained data such as vehicle positions, velocities, or trajectories, because collecting and storing such information can raise privacy issues \citep{wang2018privacy}. While many RL-ATSC approaches rely solely on aggregated traffic counts or intersection-level flow data - reducing privacy risks - some recent methods incorporate detailed trajectory data to capture finer spatiotemporal dynamics \citep{essa2020self, yazdani2023intelligent}. Although these finer-grained inputs can improve RL-ATSC results, their collection and potential sharing among intersections still require careful consideration of how to protect the information of individual drivers. Implementing data anonymization and secure data sharing protocols can help mitigate these privacy concerns while still enabling advanced RL-ATSC methods to leverage detailed traffic data.
Another potential issue in multi-agent reinforcement learning (MARL) is that sharing information between intersections can create a substantial communication burden \citep{wang2020large, wang2021adaptive}. Consequently, there are an increasing number of studies that have employed Federated Reinforcement Learning (FRL) to ATSC \citep{wang2020adaptive, fu2023federated, bao2023scalable}. 
FRL ensures vehicle privacy by keeping raw data local at each intersection and allowing each intersection to only share aggregated model parameters. It also reduces communication overhead by transmitting far less data compared to systems that require the exchange of detailed traffic information.
In FRL for ATSC, the FedAvg \citep{mcmahan2017communication} algorithm is predominantly used \citep{ye2021fedlight, fu2023federated}, allowing agents to send their model weights to the central server periodically and receive the averaged weights from other agents.

FRL has emerged as a promising approach to address privacy and communication challenges in MARL for ATSC. However, intersections in large cities often exhibit significant heterogeneity in traffic volumes, demand patterns, and road geometries. Furthermore, the mix of pedestrians, micromobility, and motor vehicles can vary significantly between different neighborhoods, reflecting various demands of traffic \citep{jing2007hierarchy}. On the supply side, traffic management strategies, such as highway ramp metering and tiered traffic signal systems, illustrate a hierarchical structure, with major and minor roads serving distinct functions in regulating the overall traffic flow \citep{han2020hierarchical}. A single, uniformly applied approach like FedAvg may struggle to account for these variations, potentially resulting in suboptimal performance in large-scale networks.

In this paper, we propose Hierarchical Federated Reinforcement Learning (HFRL) for ATSC. Our approach tackles the heterogeneity in traffic patterns, demands, and intersection layouts throughout a city by structuring learning hierarchically and sharing knowledge among intersections in a federated manner. Compared to conventional RL, which learns policies at a single intersection, and FRL, which aggregates models across distributed agents, HFRL introduces an additional hierarchical layer that coordinates local policies while still benefiting from federated updates.

\subsection{Contribution}

This is the first study to apply an HFRL framework to ATSC. To enhance existing FRL algorithms, we propose two new methods: FedFomoLight (an optimization-based approach) and FedClusterLight (a cluster-based approach). Both methods personalize local traffic light agents to accommodate the diverse conditions of different traffic environments. Our experiments on both synthetic and real-world traffic networks (New York City) show that this framework effectively groups agents according to variations in traffic demand and network topology. In particular, despite decentralized training, our approach surpasses centralized and classical FRL methods (e.g., FedAvg) in terms of travel time and waiting time. The key contributions of our work can be summarized as follows:

\begin{itemize} 

\item \textbf{Hierarchical Control:} We propose an HFRL framework for ATSC, marking the first application of hierarchical federated reinforcement learning to traffic signal control, thereby opening new avenues for adaptive traffic management. Our method addresses heterogeneity by allowing agents to form clusters or tiers that align with local traffic patterns and hierarchical road structures.

\item \textbf{High Accuracy Despite Decentralized Training:} By introducing FedFomoLight and FedClusterLight, we personalize local agents within the HFRL framework to accommodate varying traffic conditions. Our methods achieve performance comparable to centralized approaches in small synthetic networks, and surpass centralized and classical FRL methods as network heterogeneity and scale increase, especially in real-world traffic networks.

\item \textbf{Interpretable Grouping Patterns:} Our experiments reveal that the clustering and similarity metrics learned by the proposed method effectively capture patterns in both road topology and traffic demand distributions. This interpretability highlights the potential of HFRL for robust traffic control under low communication constraints. \end{itemize}





\section{RELATED WORKS}
MARL represents the state-of-the-art in ATSC systems, as it enables multiple agents to learn and coordinate optimal traffic signal settings. Currently, federated learning (FL), a distributed and collaborative framework, has proven particularly suitable for ATSC due to its ability to leverage decentralized data while preserving privacy. Recently, the integration of FL and RL has further expanded the potential of ATSC by utilizing both collaborative model training and decision-making. This section summarizes existing research on MARL for ATSC, FL-based ATSC, and various HFL approaches that address training data heterogeneity.






\subsection{MARL for Traffic Signal Control}

RL is a branch of machine learning. In ATSC, RL-based methods do not depend on traffic models, but learn from past experience and adjust traffic signal phases according to global or local environments \citep{MO2022103728, wang2024large}. MARL methods \citep{xu2021hierarchically, soleimany2023hierarchical, yu2023decentralized} surpass traditional single-agent approaches in urban ATSC by deploying individual agents at intersections to make decisions independently using local data. Typically, there are two types of MARL schemes: centralized and decentralized \citep{song2024cooperative, wei2019presslight, wang2024large}. Centralized MARL uses a central controller to oversee the entire traffic network, making decisions for all traffic signals simultaneously. In contrast, decentralized MARL allows each traffic signal to operate as an independent agent, reducing communication overhead and adapting more effectively to dynamic conditions, making it ideal for large-scale urban areas \citep{li2024cooperative, MO2022103728, zhang2024survey, aslani2017adaptive, wang2024large}.

\subsection{Federated Learning and Traffic Signal Control}

FL is a distributed collaborative learning framework that enables multiple agents to train a shared machine learning model without exchanging raw data, but by sharing models' weights. Initially introduced by Google \citep{mcmahan2017communication} to predict user text input on Android devices, FL uses local datasets on individual devices while maintaining data privacy and security. This approach is particularly advantageous in scenarios where privacy concerns and large-scale distributed systems are present, as it mitigates the need for central data pooling.

In the context of ATSC, FL offers significant improvements by enabling agents (e.g., traffic lights at different intersections) to collaborate and learn from shared experiences without requiring real-time data transfers. ATSC systems involve numerous intersections, each with unique traffic patterns, where real-time communication and the exchange of information between these agents are crucial to optimize traffic flow and reduce congestion. However, large-scale real-time communication can lead to significant overhead, which can impact the efficiency and scalability of the system. FL mitigates these challenges by enabling decentralized learning, thereby improving the scalability and robustness of MARL-based ATSC approaches.

Tab.~\ref{tab:research_summary} summarizes recent applications of FL frameworks integrated with MARL for TSC, highlighting key contributions, RL algorithms, and associated frameworks. These approaches aim to improve traffic control by leveraging both FL's privacy-preserving benefits and the decentralized coordination of agents in complex traffic environments.

\begin{table}[h]
\caption{Summary of papers about FL and ATSC}
\label{tab:research_summary}
{\small
\begin{tabular}{p{1.5cm}|>{\raggedright\arraybackslash}p{9.3cm}|>{\raggedright\arraybackslash}p{1.3cm}|>{\raggedright\arraybackslash}p{2.5cm}}

\hline
Ref. & Key Contributions & RL Algorithms &  Framework  \\ 

\hline
\hline

\cite{wang2020adaptive} & Introduces two MARL control modes and a FRL framework to solve the ATSC problem. It focuses on improving robustness and computational efficiency by distributing control across multiple agents, allowing local optimization at each intersection. & A2C & MARL with independent and joint control mode and FL\\ 
\hline
\cite{ye2021fedlight} & Presents a cloud-based FRL framework that coordinates decentralized agent training for optimizing multi-intersection traffic control. Uses FL-like gradient aggregation and parameter sharing to enhance training convergence and enables knowledge sharing among agents. & A2C & MARL with FL  \\ 
\hline

\cite{hudson2022smart} & Introduces a SEAL framework to represent traffic conditions across diverse intersections and proposes a FRL approach for training traffic light decision-making policies. Improves the reward-communication trade-off, reducing communication costs by up to 36.24\% while maintaining only a 2.11\% performance decrease compared to centralized training. & PPO & MARL with FL  \\ 
\hline
\cite{fu2023federated} & Proposes an open-source simulation framework that integrates SUMO-Gym with Ray's Rllib for FRL to solve the ATSC problem. Introduces the first A3C-based FL framework for ATSC and tests the approach on real-world road networks, calibrating traffic volumes to analyze communication costs. & A3C & MARL with FL  \\
\hline
\cite{soleimany2023hierarchical} & Introduces a hierarchical learning model to address scalability in traffic signal management. Uses profilization-based learning with FRL to process data locally at node and intersection levels, optimizing traffic flow while maintaining source-to-destination delays (SDDs) with deep RL. & DRL & RL on intersection level with FL on cloud level  \\
\hline
\cite{bao2023scalable} & Proposes an FRL method for TSC. Introduces a unified state representation and action selection method to handle diverse intersection structures, using partial aggregation and fine-tuning to preserve local specificity. & DQN & Deep Reinforcement Learning with FL \\ 
\hline
\hline
\end{tabular}%
}
\end{table}

\subsection{Heterogeneous Federated Learning}

Existing FL approaches predominantly assume that all clients share a uniform network structure or similar data distributions. However, practical scenarios often exhibit substantial differences among clients in terms of data distributions, model structures, communication networks, and edge devices \citep{ye2023heterogeneous}. 
Hierarchy naturally arises due to asymmetry in demand and supply. 
For instance, intersections near schools may experience different peak traffic hours compared to those near sports arenas. To address FL heterogeneity, several state-of-the-art methods have been proposed. These methods, known as HFL, are categorized into data-level, model-level, and server-level approaches\citep{ye2023heterogeneous}. Among these approaches, it is crucial to identify one that can account for the heterogeneity of different intersections by dynamically grouping similar ones and training models that reflect their specific characteristics.

\subsubsection{Data-level Methods}

Data-level methods focus primarily on tasks related to data management, including data augmentation and anonymization techniques. Significant contributions in this area include \citep{li2020federated} \citep{li2019fedmd} \citep{yoon2021fedmix}. While these efforts are aimed at mitigating statistical heterogeneity between clients and enhancing data privacy, they may not be ideally suited for grouping and training models that accurately reflect the unique characteristics of diverse datasets.

\subsubsection{Model-level Methods}

Model-level methods focus on operations designed specifically for the model layer, including sharing partial structures and model optimization. Recent research on $\Psi$-Net \citep{yu2022heterogeneous} and QAvg $\&$ PAvg \citep{jin2022federated} optimizes policy performance in diverse environments without data sharing, although mainly for tabular methods. Per-FedAvg \citep{fallah2020personalized} and FedPer \citep{arivazhagan2019federated} focus on personalized federated learning, primarily for supervised learning environments. Related work in FedFormer \citep{hebert2022fedformer} and the KT-pFL framework \citep{zhang2021parameterized} enhances personalization through innovative model structures but with potential limitations in direct policy updates.

\subsubsection{Server-level Methods}

Server-level methods focus on the central server's engagement, including tasks such as client selection, grouping, and coordination. Related works include deep reinforcement learning (DRL) on the heterogeneous training intensity assignment
problem for FL \citep{9965742} and RL client grouping \citep{9134408}, where both focus on dynamic adjustment via RL. Other works include FedCor \citep{tang2022fedcor} and QAvg $\&$ PAvg \citep{jin2022federated}. 

Cluster-based and optimization-based approaches further enhance client coordination and model personalization at the server-level. Cluster-based methods infer the heterogeneity of the data distribution using model information, assigning clients to clusters for model aggregation. Examples include research on FedCHAR $\&$ FedCHAR-DC \citep{li2023hierarchical}, the Federated Multitask Learning (FMTL) framework \citep{sattler2020clustered}, and the Iterative Federated Clustering Algorithm (IFCA) \citep{ghosh2020efficient}. Other clustering methods, such as hierarchical clustering~\citep{briggs2020federated}, EM clustering~\citep{long2023multi}, and KMeans~\citep{ghosh2019robust}, help FL converge faster with lower communication costs. In optimization-based approaches, client aggregation is often formulated as an integer programming problem to minimize objective functions. Such methods include Ditto \citep{li2021ditto}, the APPLE framework \citep{luo2022adapt}, and FedFomo \citep{zhang2020personalized}.

Typical HFL methods rely on static groupings of clients, assuming that their clusters remain fixed over time. However, in real-world ATSC scenarios, traffic conditions evolve, as do the interactions among roads and intersections. To address this gap, we propose a novel HFRL framework that learns to group road interactions dynamically based on ambient conditions. Specifically, we use clustering-based or optimization-based strategies to perform FedAvg independently within groups of intersections that share similar characteristics, while allowing these groups to adapt as traffic environments change. The resulting algorithms, FedFomoLight and FedClusterLight, dynamically adapt to evolving traffic environments.

The remainder of the paper is organized as follows. Sec.~\ref{PS} introduces the definition of the ATSC problem in MARL. In Sec.~\ref{sec:methodology}, we detail the proposed framework and algorithms. Sec.~\ref{sec:exp} describes the experimental setup and presents both performance evaluations and a sensitivity analysis. Finally, Sec.~\ref{conclusion} concludes this study.

\begingroup
\setlength{\tabcolsep}{6pt}
\renewcommand{\arraystretch}{1.5}
\begin{table}[H]
\begin{tabular}{  m{3cm}  m{12cm} } 
  \hline
  Notations & Definitions \\ 
  \hline
  \hline
   $t$ & FL server training round \\ 
  \hline
   $\tau$ & local training round\\ 
  \hline
   $s$ & state \\ 
  \hline
   $S$ & state space\\ 
  \hline
  $a$ & action \\ 
  \hline
  $A$ & action space \\ 
  \hline
  $o$ & observation \\ 
  \hline
  $O$ & observation space \\ 
  \hline
  $o_n$ & lane occupancy for intersection $n$ \\ 
  \hline
  $q_n$ & queue length \\ 
  \hline
  $\overline{v}_n$ & average speed for intersection $n$ \\ 
  \hline
  $ph_n$ & ratio of the total phase length of each signal light type for intersection $n$ \\ 
  \hline
  $\gamma$ & discounted factor \\ 
  \hline
  $\pi$ & policy \\ 
  \hline
  $N$ & number of agents \\ 
  \hline
  $n$ & client $n$ in FL\\
  \hline
  $K$ & total number of clusters\\
  \hline
  $C$ & set of clusters\\
  \hline
  $\delta$ & TD error\\
  \hline
  $\alpha$ & learning rate\\
  \hline
  $A(s, a)$ & advantage function\\
  \hline
  $\phi$ & parameters of critic network\\
  \hline
  $\theta$ & parameters of actor network\\
  \hline
  $w$ & set of network parameters\\
  \hline
  $\mathcal L$ & FedFomo optimization objectives\\
  \hline
  $\rho$ & weights of candidate federated models\\
  \hline
  $V_{\phi}(s)$ & value function\\
  \hline
  $\pi_\theta$ & current policy\\
  \hline
  $D$ & global update rounds\\

  \hline
  \hline
\end{tabular}
\caption{Lists of Notations}\label{label:RL_table}
\end{table}
\endgroup

\section{PROBLEM STATEMENT}
\label{PS}
\begin{figure}[H]
  \centering
  \includegraphics[width=1.0\linewidth]{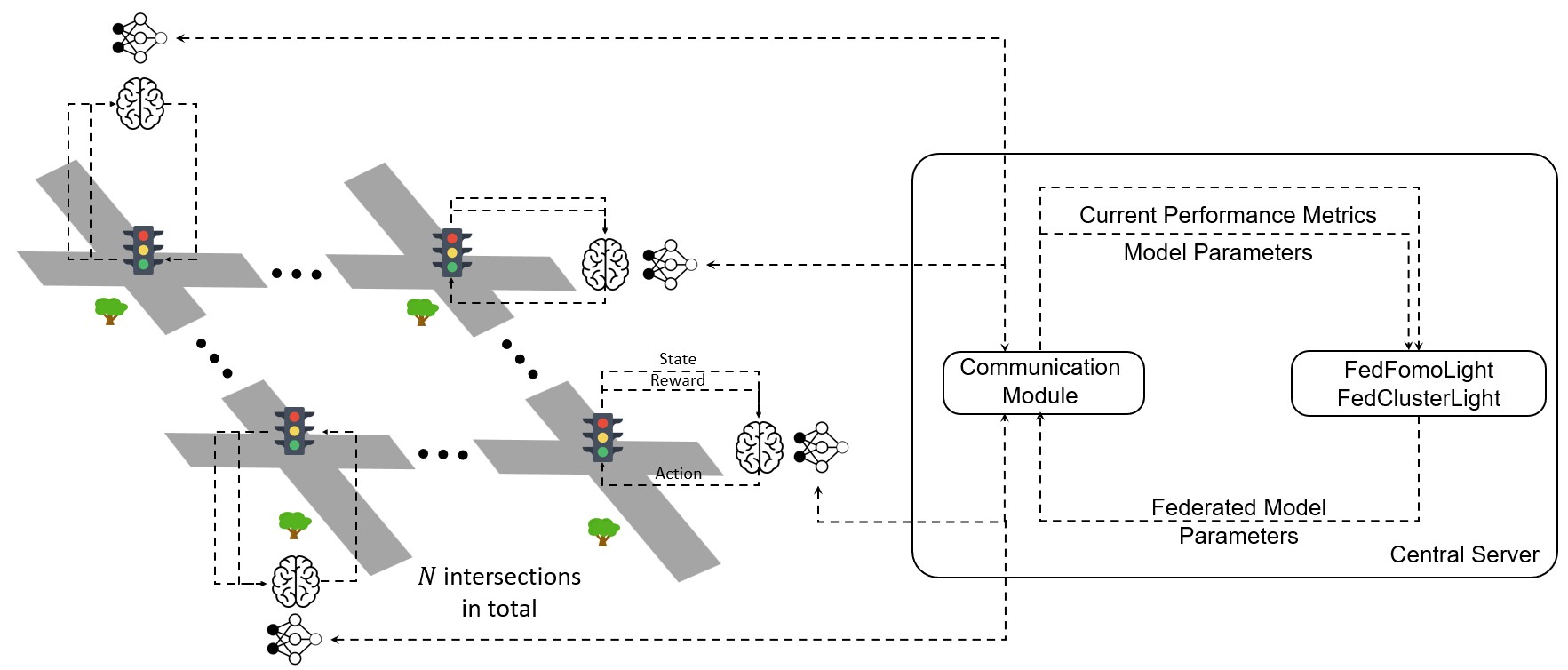}
  \caption{The proposed HFRL framework for ATSC.}
  \label{fig:ATSC-and-HFL}
\end{figure}



Fig.~\ref{fig:ATSC-and-HFL} illustrates the proposed HFRL framework for ATSC. The left side represents the decentralized ATSC architecture, where traffic signals are managed independently by local agents. The right side highlights the hierarchical learning process implemented on a central server, which coordinates knowledge sharing across agents. A detailed explanation of this methodology is provided in Section~\ref{sec:methodology}.

Fig.~\ref{fig:ATSC-and-HFL} presents ATSC as a dynamic multi-agent traffic management strategy that optimizes signal timings in real time to alleviate congestion. In our configuration, an ATSC system is supported by edge computing devices and a central server, over a road traffic network with $N$ nodes. Urban road networks exhibit diverse and complex traffic conditions at multiple scales. A hierarchical approach to ATSC enables intersection-level strategies to handle localized dynamics while ensuring network-wide coordination.

\subsection{MARL}

MARL is a widely used approach for ATSC, as depicted on the left side of Fig.~\ref{fig:ATSC-and-HFL}. 
MARL frameworks can be either distributed or centralized. In this work, we adopt a distributed framework where each traffic signal is controlled by an independent RL agent trained locally. Consequently, each traffic signal is managed by an individual RL agent that observes local traffic conditions (for example, queue lengths, speeds, and signal phases) and continually updates its policy based on a locally received reward. By operating in parallel and learning simultaneously, these agents can collectively optimize traffic flow across the network while requiring only limited communication overhead.

Formally, the ATSC on a road network problem is modeled as a Markov game defined by the tuple $\bigl(N, S, O, A, P, R, \gamma\bigr)$, where $N$ is the number of agents (intersections), and $S, O, A, P, R, \gamma$ denote the state space, joint observation space, action space, state transition functions, reward functions, and discount factor, respectively. Since each agent has only partially observability of the global state, the Markov game is formulated as a partially observable Markov decision process (POMDP). At each time step, each agent chooses an action based on its local policy, transitioning the environment to a new state, and generating a local reward. This cycle repeats until a terminal condition is reached, and agents refine their policies to maximize their individual rewards. However, such self-interested optimization may not directly lead to improved network-wide performance, which motivates our proposed HFRL framework in Sec.~\ref{frl-a2c}.

We define the state space, the action space, and the local reward function for each traffic light agent as follows:

\begin{enumerate}
    \item \textbf{State ($S$).}
    For a traffic light $n$ at round $\tau$, the local state includes lane occupancy $\bigl(o_n^\tau\bigr)$, queue length $\bigl(q_n^\tau\bigr)$, average speed $\bigl(\overline{u}_n^\tau\bigr)$, and the phase ratios $\bigl(ph_n^\tau\bigr)$ for the relevant signal phases (e.g., green, yellow, red). We follow part of the settings in Hudson's work~\citep{hudson2022smart}. Although traffic variables are aggregated across lanes, the inclusion of phase ratios preserves information about which lane groups are currently being served, allowing the agent to distinguish traffic conditions across different active phases.

\begin{table}[ht]

\centering
\caption{Summary of local state features}
\label{tab:state_features}
\begin{tabular}{p{2.2cm} p{5.5cm} p{0.8cm} p{2.6cm} p{2.6cm}}
\hline
\textbf{Feature} & \textbf{Definition} & \textbf{Unit} & \textbf{Normalization} & \textbf{Aggregation Level} \\
\hline
Lane Occupancy ($o_n^\tau$) 
& Ratio of lane length occupied by vehicles across lanes controlled by traffic light $n$. 
& -- 
& $[0,1]$ (inherently normalized) 
& Per signal (averaged over lanes) \\
\hline
Queue Length ($q_n^\tau$) 
& Ratio of lane length occupied by halted vehicles (speed $\leq 0.1$ m/s) across lanes controlled by traffic light $n$. 
& -- 
& $[0,1]$ (inherently normalized) 
& Per signal (averaged over lanes) \\
\hline
Average Speed ($\overline{u}_n^\tau$) 
& Average vehicle speed on lanes controlled by traffic light $n$. 
& m/s 
& Divided by road speed limit to $[0,1]$ 
& Per signal (aggregated across lanes) \\
\hline
Phase Ratios ($ph_n^\tau$) 
& \textbf{Current-phase encoding}. Ratio of signal colors (green, yellow, red) in the currently active phase of traffic light $n$. 
& -- 
& $[0,1]$ (per phase state) 
& Per signal (current phase) \\
\hline
\end{tabular}

\end{table}

    \item \textbf{Action ($A$).}
    To ensure consistency across agents in the federated learning setting, we define a binary action space $a_n^\tau \in \{0,1\}$. Here, $a_n^\tau = 1$ indicates that traffic light $n$ will attempt to switch to the next phase at round $\tau$, while $a_n^\tau = 0$ means that it remains in its current phase.  The phase sequence itself follows a fixed, pre-defined transition order that ensures non-conflicting traffic flows. When the agent selects the switch action, the system advances to the next phase in this preset cycle (e.g., green → yellow → red → green), subject to minimum and maximum timing constraints between transitions \citep{hudson2022smart}.

    \item \textbf{Local Reward ($R$).}
    In a distributed framework, each traffic light agent receives an individual reward to reduce local congestion:
    \begin{equation}
    R_n^\tau = -\bigl(o_n^\tau + q_n^\tau\bigr)^2,
    \end{equation}
    where higher occupancy or halted queues at intersection $n$ lead to a more negative reward, encouraging the agent to alleviate congestion locally.
\end{enumerate}

In RL, each agent additionally learns a value function \(V_{\phi}(s)\) that estimates the expected discounted return when starting from state \(s\) under the current policy \(\pi_\theta\):
\begin{equation}
V(s^\tau) 
= \mathbb{E}_{l \sim \pi} \Bigl[\sum_{m=0}^{\infty} \gamma^m \, R^{\tau+m} \;\Big|\; s^\tau\Bigr].
\end{equation}
where $l$ denotes a trajectory generated by following the policy $\pi$.This value function serves as a baseline to reduce variance in policy updates and helps guide each agent toward actions that improve long-term performance.

The ATSC system follows a hierarchical MARL framework. At the lower level, each traffic signal agent directly interacts with vehicles at its intersection, making real-time decisions based on local traffic conditions. Edge device, positioned at the same level, collect intersection-specific data and train local models before sharing updates with the central server. These edge devices then communicate their updates to a central server. At the upper level, the central server aggregates model parameters received from all edge devices, ensuring consistency and optimizing global traffic flow throughout the entire network. Consequently, the intersection-level agents manage local traffic interactions, the edge devices handle localized modeling and training, and the central server coordinates global optimization efforts. This structure effectively combines decentralized decision-making with centralized coordination for efficient urban traffic management.

\subsection{FRL-A2C}
\label{frl-a2c}

MARL can be adapted to the federated learning to enhance privacy, reduce communication overhead, and facilitate coordinated learning among distributed agents. In this work, we employ synchronized FRL, where model weights are aggregated synchronously following local RL training. In the context of ATSC, MARL is responsible for optimizing the policy of each local agent using only local observations and rewards. By interacting with vehicles in its immediate vicinity, each traffic signal agent refines its control strategy (e.g., deciding when to switch phases) to maximize long-term performance measures such as reduced queue length or minimized travel delay. Meanwhile, FL coordinates knowledge sharing among these decentralized RL agents. Instead of transmitting raw observation data, agents exchange only their locally-trained model parameters with the central FL server. This arrangement reduces communication overhead and preserves data privacy, while still allowing each agent to benefit from updates derived from other intersections. As a result, FRL leverages both local exploration (through RL) and global knowledge aggregation (through FL), enabling more robust and scalable solutions for multi-agent traffic signal control.




We employ the Advantage Actor-Critic (A2C) approach in our federated RL framework. In each local training round $\tau$, each agent \(n\) maintains a local model \(w_n\), consisting of its Actor network parameters \(\theta_n\) and Critic network parameters \(\phi_n\):
\begin{equation}
w^\tau_n
\;=\;
\bigl(\theta^\tau_n,\; \phi^\tau_n\bigr).
\end{equation}

In an FRL setup, a central FL server coordinates the training process among multiple clients (agents). Each client maintains a local model and interacts with its own environment to collect experience. The central FL server initializes a global model and distributes it to clients. Once clients receive this global model, they proceed to perform local RL updates based on their private data or environmental interactions.
\subsubsection{Client Side}


In FRL, each client $n \in \mathcal{N} = \{1, \dots, N\}$ corresponds to a traffic light with local model parameters $w_n^\tau$ in the local round $\tau \in \{1, \dots, \mathcal{T}\}$, 
where $\mathcal{T}$ denotes the total number of local interaction time steps within one communication round.
 At the start of each communication round, a central FL server distributes the current global model to a subset of these clients. Each client then trains its local model by optimizing a local reinforcement learning objective \(\mathcal{R}_n(w_n^\tau)\), using data gathered from its own environment. Once training is complete, each client uploads its updated parameters to the server. The server aggregates these updates and redistributes the resulting global model for the next round. This process facilitates collaborative learning across multiple agents, reduces communication overhead, and preserves data privacy. In its simplest form, each agent’s parameters may be updated as follows:
\begin{equation}
w_n^{\tau+1} \;\leftarrow\; 
w_n^\tau \;-\; \eta \,\nabla R_n\bigl(w_n^\tau\bigr),
\end{equation}

where \(\eta\) is the learning rate, and \(\nabla R_n(\cdot)\) denotes the gradient based on the agent’s local observations. In practice, the updates for each agent proceed as follows:

\paragraph{Environment Interaction and Trajectory Collection:}
Each agent interacts with its local environment for $\mathcal{T}$ rounds. During this phase, it collects trajectories of states, actions, and rewards:

\begin{equation}
\left\{ \left(s_n^{\tau},\, a_n^{\tau},\, R_n^{\tau},\, s_n^{\tau+1}\right) \right\}_{\tau=1}^{\cal T}.
\end{equation}
Here, \(s_n^{\tau}\) is the traffic state of agent \(n\) at round \(\tau\), \(a_n^{\tau}\) is the selected action, and \(R_n^{\tau}\) is the corresponding reward.
\paragraph{Value Target and Advantage Computation:}
To compute the advantage function \(A(s_n^{\tau},\, a_n^{\tau})\), A2C typically uses a \(\cal K\)-step return or a single-step temporal difference (TD) error. A common approach is:
\[
A\bigl(s_n^{\tau}, a_n^{\tau}\bigr) \;=\; 
\sum_{i=0}^{{\cal K}-1} \gamma^i \, R_n^{\tau+i}
\;+\; \gamma^{\cal K} \, V_{\phi_n}\bigl(s_n^{\tau+{\cal K}}\bigr)
\;-\; V_{\phi_n}\bigl(s_n^{\tau}\bigr),
\]
where \(\gamma\) is the discount factor, and \(V_{\phi_n}(\cdot)\) is the value function parameterized by \(\phi_n\).

\paragraph{Critic Network Update (value function parameters \(\phi_n\)):}
\[
\phi_{n}^{\tau+1} \;\leftarrow\; \phi_{n}^{\tau} 
\;+\; \eta\, A\bigl(s_{n}^{\tau}, a_{n}^{\tau}\bigr) 
\,\nabla_{\phi} V_{\phi}\bigl(s_{n}^{\tau}\bigr),
\]
where \(A\bigl(s_{n}^{\tau}, a_{n}^{\tau}\bigr)\) is the advantage function, and \(V_{\phi}\bigl(s_{n}^{\tau}\bigr)\) is the value function.

\paragraph{Actor Network Update (policy parameters \(\theta_n\)):}
\[
\theta_{n}^{\tau+1} \;\leftarrow\; \theta_{n}^{\tau} 
\;+\; \eta\, 
\nabla_{\theta} \log \pi_{\theta}\bigl(a_{n}^{\tau}\mid s_{n}^{\tau}\bigr)\,
A\bigl(s_{n}^{\tau}, a_{n}^{\tau}\bigr).
\]
where $\pi_{\theta}$ is the policy function.

By iterating between these local A2C updates and global aggregations, the FRL framework enables a scalable and privacy-preserving approach to decentralized traffic light control.

\subsubsection{Server Side}
 Once the local training of the round is complete, each agent \(n\) sends its updated parameters \(w_n^{t}\) of global round $t$ to the central FL server. In the following text, the superscript of $w$ denotes the round of global updates. The server then aggregates these parameters using the FedAvg rule:
\begin{equation}
w^{t} 
\;\leftarrow\; \frac{1}{N} \sum_{n=1}^{N} w_n^{t}.
\end{equation}
The newly averaged model \(w^{t}\) is then broadcast to the agents for the next round of local training, thereby iterating between local reinforcement learning and global model aggregation.

Although FedAvg is straightforward and communication efficient, it can struggle in complex, heterogeneous traffic networks. Simply averaging local models can reduce the ability to adapt to diverse intersection conditions and dynamic traffic flows, limiting the method’s overall generalizability. Consequently, advanced HFRL frameworks have been proposed to address these shortcomings by introducing additional structural elements that better capture heterogeneity and enhance the overall learning performance in multi-agent ATSC. We provide detailed descriptions of the two proposed HFRL algorithms in the following section.

\section{METHODOLOGY}
\label{sec:methodology}

Urban road networks inherently have hierarchical structures due to varying roles and traffic flow patterns across different intersections and road types. To effectively address this hierarchical nature, our proposed framework incorporates either personalized or cluster-based aggregation of local model parameters, explicitly accounting for model similarities and local performance.

\subsection{Hierarchical FRL Algorithms}

\begin{figure}[H]
  \centering
  \includegraphics[width=1.0\linewidth]{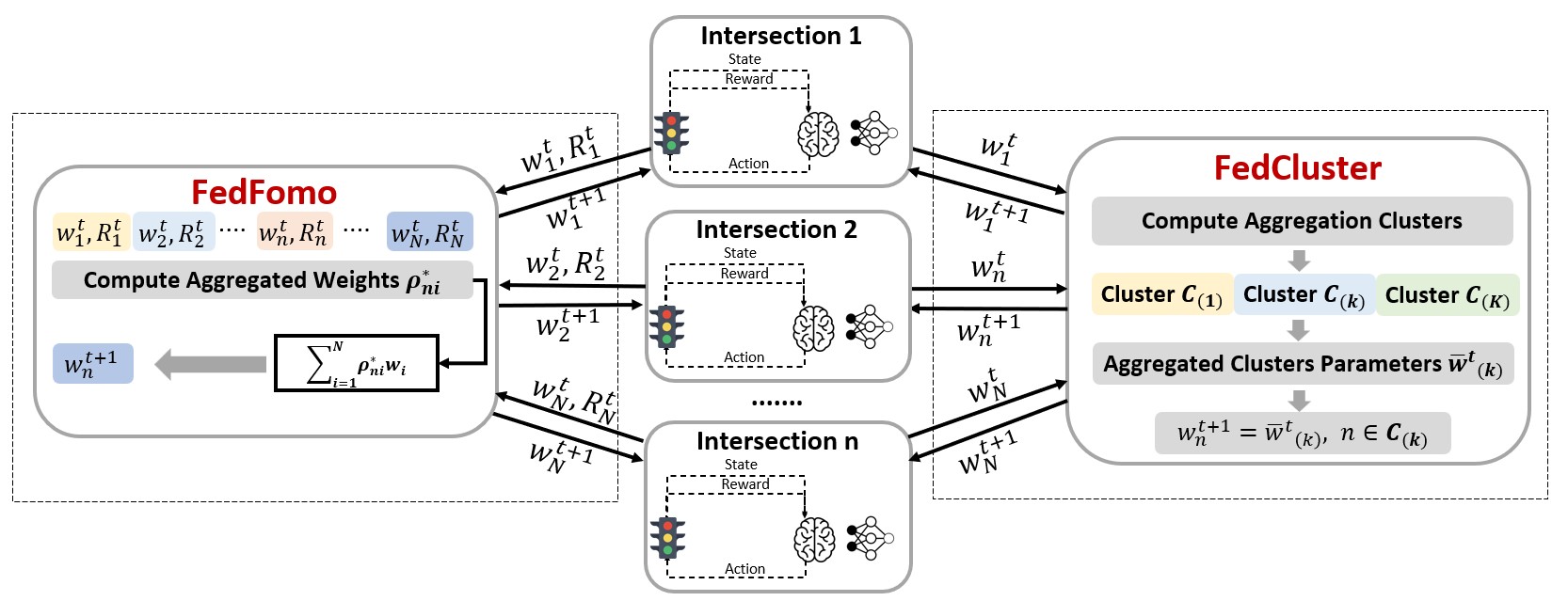}
  \caption{Illustration of the proposed HFRL algorithms (FedFomoLight and FedClusterLight) within the dashed-line box, with one method applied during execution}
  \label{fig:Fed-cluster-fomo}
\end{figure}

In the hierarchical FRL framework, each local model is trained within a MARL environment using only its local observations. Following a predefined number of local training rounds, a communication round begins during which each local model uploads its learned parameters and corresponding rewards to a central FL server. The server then applies one of two aggregation methods, as illustrated in Fig.~\ref{fig:Fed-cluster-fomo}:

\begin{itemize}
\item \textbf{FedFomoLight:} An optimization-based aggregation method where each local participant receives a personalized aggregated model. This model is generated using a weighted combination of other local models, where the weights depend on local performance rewards and model similarity.
\item \textbf{FedClusterLight:} A clustering-based aggregation method where local models are first grouped into clusters. Then aggregation occurs within these clusters, allowing participants with similar traffic characteristics to benefit collectively.
\end{itemize}

After aggregation, each local model receives its updated federated model from the central FL server and proceeds with local training until convergence. In addition, both FedFomoLight and FedClusterLight evolve dynamically during the RL training process, allowing group assignments and aggregation strategies to adapt in response to the changing behaviors and performance of individual agents. Below, we focus primarily on server-side algorithms, specifically detailing our modifications to FRL, while assuming that all other configurations remain consistent between FedFomoLight and FedClusterLight.

\subsection{FedFomoLight}
\label{sec:FedFomoLight-A2C}

We adapt the general FedFomo framework to MARL, resulting in the proposed FedFomoLight. Each client $n$ maintains a personalized objective aimed at maximizing its objective function $\mathcal{L}_n(w^t_n)$ given the current local model parameters $w^t_n$ in round $t$. The global optimization objective is then defined as finding the optimal set of local model parameters:

\begin{equation}
    \mathcal{L}_n(w^t_n) = A\bigl(s_{n}^{t}, a_{n}^{t}\bigr) 
\,\nabla_{\phi} V_{\phi}\bigl(s_{n}^{t}\bigr) + \nabla_{\theta} \log \pi_{\theta}\bigl(a_{n}^{t}\mid s_{n}^{t}\bigr)\,
A\bigl(s_{n}^{t}, a_{n}^{t}\bigr)
\end{equation}

\begin{equation}
\bigl\{(w^t_1)^*, (w^t_2)^*\dots, (w^t_{|N|})^*\bigr\}
\;=\;
\arg\!\min_{\{w^t_n\}_{n=1}^{|N|}}
\;\sum_{n=1}^{|N|}
\mathcal{L}_n(w^t_n),
\end{equation}

Thus, each agent strives to maximize its personalized RL objective $\mathcal{L}_n$, while still benefiting from collaborative updates provided by the federated framework. Formally, this MARL-adapted formulation parallels the supervised definition in the original FedFomo framework, but incorporates RL-specific objectives $\mathcal{L}_n$.

Consequently, FedFomoLight offers enhanced flexibility and personalized adaptation suitable for heterogeneous traffic scenarios compared to conventional FedAvg-style methods, without requiring the exchange of raw experience data among agents. A comprehensive overview of the proposed algorithm is provided in Alg.~\ref{alg:FedFomo_RL}.

\paragraph{\textbf{Weight Optimization:}}

Unlike FedAvg, which assigns equal weight to all models (or weights them by local dataset size), for each agent $n$ FedFomoLight optimizes personalized importance \(\boldsymbol{\rho_n} = [\rho_{n1},\dots,\rho_{nN}]\), where \(\sum_{i=1}^N \rho_{ni}=1\) to improve the local RL objective \(\mathcal{L}_n\). According to Equ.~\eqref{equ:fedfomo_update_first}, the model deltas $w_i^{t} - w_n^{t-1}$ determine how much each client should adjust its local model parameters to optimize for \(\mathcal{L}_n\). The steps we can take toward this objective are restricted by the fixed model parameters $w_i^{t}$ available at time $t$. For personalized or non-IID target distributions, we can iteratively solve for the optimal combination of client models using Equ.~\eqref{equ: weight_update} after the deviation from $t$ to $t+1$ \citep{xiong2024personalized}.

\begin{equation}
\label{equ:fedfomo_update_first}
w_n^{t+1}
\;\leftarrow\;
w_n^{t-1}
\;+\;
\sum_{i=1}^{N} 
\rho_{ni} 
\;\bigl(w_i^{t} - w_n^{t-1}\bigr),
\end{equation}

\begin{equation}
\label{equ: weight_update}
w_n^{t+1}
\;\leftarrow\;
w_n^{t-1} 
\;-\;
\alpha\, 
\mathbf{1}^T \nabla_{\boldsymbol{\rho_n}}
\;\mathcal{L}_n\bigl(w_n^{t-1}\bigr),
\end{equation}

where \(\alpha\) is a learning rate. By combining the updating method of model weights prior to the current state in Equ.~\eqref{equ:fedfomo_update_first} and Equ.~\eqref{equ: weight_update}, we obtain:
\begin{equation}
\label{equ: two_equations}
\sum_{i=1}^{N} 
\rho_{ni} 
\;\bigl(w_i^{t} - w_n^{t-1}\bigr)
\;=\;
-\,\alpha\;\mathbf{1}^T
\nabla_{\boldsymbol{\rho_n}}\;
\mathcal{L}_n\bigl(w_n^{t-1}\bigr).
\end{equation}

To approximate \(\frac{\partial}{\partial \rho_{ni}}\,\mathcal{L}_n(w_n^{t-1})\),
we employ a first-order Taylor expansion. Define:
\begin{equation}
\label{equ: A2_equations}
\varphi_{ni}(\rho_{ni})
\;=\;
\rho_{ni}\,w_i^{t}
\;+\;
(1-\rho_{ni})\,w_n^{t-1}.
\end{equation}
Evaluating \(\mathcal{L}_n(\varphi_{ni}(\rho_{ni}))\) near \(\rho_{ni}=0\) and then setting \(\rho_{ni}=1\) provides:
\begin{equation}
\label{equ: taylor_equations_1}
\frac{\partial}{\partial \rho_{ni}}\,\mathcal{L}_n\bigl(w_n^{t-1}\bigr)
\;\approx\;
\mathcal{L}_n\bigl(w_i^{t}\bigr)
\;-\;
\mathcal{L}_n\bigl(w_n^{t-1}\bigr).
\end{equation}
Substituting back into Equ.~\eqref{equ: two_equations}, we solve for \(\rho_{ni}\):
\begin{equation}
\label{equ: final_weights_cal}
\rho_{ni}
\;=\;
-\,\alpha\,
\frac{ 
  \mathcal{L}_n\bigl(w_i^{t}\bigr)
  \;-\;
  \mathcal{L}_n\bigl(w_n^{t-1}\bigr)
}{
  \bigl\|w_i^{t} - w_n^{t-1}\bigr\|
}.
\end{equation}
Finally, we normalize each \(\rho_{ni}^*\) to ensure non-negativity and that \(\sum_i \rho_{ni}^*=1\):
\begin{equation}
\rho_{ni}^*
\;=\;
\frac{\max(\rho_{ni}, 0)}{\sum_{i}\max(\rho_{ni}, 0)}.
\end{equation}

\paragraph{\textbf{Weight Updates:}}

FedFomoLight learns a set of personalized importance \(\{\rho^*_{ni}\}\) that control how each agent \(i\) combines the model parameters of other agents \(n\). Specifically, given a set of available federated model candidates \(\{w_i^{t}\}\), agent \(n\) updates its own model \(w_n^{t+1}\) as:
\begin{equation}
\label{equ: fedfomo_update}
w_n^{t+1}
\;\leftarrow\;
w_n^{t-1}
\;+\;
\sum_{i=1}^{N} 
\rho^*_{ni} 
\;\bigl(w_i^{t} - w_n^{t-1}\bigr),
\end{equation}
where \(t\) indexes the iteration or communication round.
Because the same importance scheme \(\{\rho_{ni}^*\}\) applies to both \(\theta_n^t\) and \(\phi_n^t\), FedFomoLight naturally handles the combined actor-critic model in a single federated update step.

By embedding the FedFomo objective into a multi-agent RL framework, FedFomoLight allows each local agent to tailor its Actor-Critic parameters to local conditions while still benefiting from shared knowledge.

\begin{algorithm}[H]
\begin{algorithmic}[1]

\STATE \textbf{Initialize:} For each agent \(n \in \{1,\dots,N\}\), initialize local parameters $w_i^0$

\STATE \textbf{Set:} Number of global rounds \(T\), local training steps per round \(\cal T\).

\FOR{each global round \(t = 1, 2, \dots, T\)}
    \STATE \textbf{Local A2C Training:}
    \FOR{each agent \(n\) in parallel}
        \STATE Run A2C locally for \(\cal T\) timesteps using \(w_n^{t-1}\).
        \STATE Obtain updated local parameters \(w_n^t\).
        \STATE Upload \(w_n^t\) to the central FL server.
    \ENDFOR

    \STATE \textbf{Importance Computation:}
    \STATE Collect \(\{w_1^t, \dots, w_N^r\}\).
    \FOR{each agent \(n\)}
        \STATE Let \(\boldsymbol{\rho_n} = [\rho_{n1},\dots,\rho_{nN}]\) be the importance vector for agent \(n\).
        \FOR{each agent \(n\)}
            \STATE Compute raw importance \(\rho_{ni}\) via Equ.~\eqref{equ: final_weights_cal}
        \ENDFOR
        \STATE \textbf{Normalize} to ensure \(\rho_{ni} \ge 0\) and \(\sum_i \rho_{ni} = 1\)
    \ENDFOR

    \STATE \textbf{Broadcast to Clusters:}
    \FOR{each agent \(n\)}
        \STATE Update model parameters based on Equ.~\eqref{equ: fedfomo_update}.
        \STATE Return updated \(w_n^t\) to agent \(n\).
    \ENDFOR
\ENDFOR

\end{algorithmic}
\caption{FedFomoLight}
\label{alg:FedFomo_RL}
\end{algorithm}

\subsection{FedClusterLight}

We also apply a clustering-based approach, FedClusterLight, to address this problem, as illustrated on the left-hand side of Fig.~\ref{fig:Fed-cluster-fomo}. In round $t$, each client $n$ has local model parameters $w_i^{t}$. Although standard FRL typically assumes a single global reward or loss function, clustering-based approaches recognize that each client may belong to a subset of agents that share similar characteristics. We detail each step of FedClusterLight in the following section and in Alg.~\ref{alg:FedRL_clustering}.

\paragraph{\textbf{Clustering-Based Federation}}
FedClusterLight builds on the idea that agents with similar model attributes, hyperparameters, or traffic patterns can be grouped together. After each local training session, a central FL server partitions agents into $K$ clusters $\{C_{(1)},\dots, C_{(K)}\}$ through a clustering algorithm such as K-means:
\begin{equation}
\{C_{(1)}, \dots, C_{(K)}\}
\;=\;
\mathrm{Kmeans}\Bigl(\{w_1^{t},\dots,w_{|N|}^{t}\}\Bigr).
\label{equ:clusterlight_cluster_step}
\end{equation}
Each cluster $C_{(k)}$ contains a subset of agents selected from the full set \{1, \dots, N\} that exhibit similarity in their parameters or other chosen attributes (for instance, traffic density levels, local observations, or state-action distributions).


\paragraph{\textbf{Within-Cluster Aggregation}}
After local training, each agent $n$ uploads its model $w_n^t$ to the server, which performs clustering (Eq.~\eqref{equ:clusterlight_cluster_step}). Once clusters are formed, each cluster $C_{(k)}$ executes a FedAvg-like update on its members:

\begin{equation}
\qquad
\overline{w}^t_{(k)}
\;=\;
\frac{1}{|C_{(k)}|}
\sum_{i\in C_{(k)}}w_i^t,
\label{equ:clusterlight_avg}
\end{equation}
yielding aggregated Actor and Critic parameters $\overline{w}^t_{(k)}$ for cluster $k$. This “cluster-level” model reflects the shared traffic characteristics of $C_{(k)}$.

\paragraph{\textbf{Broadcast Cluster-Specific Updates}}
Finally, each agent $n\in C_{(k)}$ receives the aggregated parameters:

\begin{equation}
w_n^{t+1}\leftarrow\;
\overline{w}^t_{(k)},\,
\label{equ:clusterlight_broadcast}
\end{equation}
and uses them as initialization for the next local training phase. This clustering-based process repeats for a specified number of global rounds $R$. In effect, agents within the same cluster share a common baseline model, while still adapting to local nuances through additional local A2C steps.

Unlike FedFomoLight, which assigns personalized weights to each foreign model, FedClusterLight operates at higher granularity by forming clusters of agents with similar attributes. All agents in a cluster receive the same aggregated parameters, simplifying computation but potentially forfeiting the finer personalization offered by FedFomoLight. Nonetheless, by aligning agents with broadly similar traffic conditions, FedClusterLight can be effective in large-scale scenarios where a purely personalized scheme may be overly complex or expensive to compute. Agents with similar phase structures, traffic demands, or parameter configurations naturally cluster together, improving learning stability and performance in heterogeneous traffic networks. Meanwhile, the method retains the communication benefits of traditional FedAvg, ensuring scalability for large multi-agent systems.


\begin{algorithm}[H]
\caption{FedClusterLight}
\label{alg:FedRL_clustering}
\begin{algorithmic}[1]
\STATE \textbf{Initialize:} For each agent \( n \in \{1,\dots,N\}\), initialize local Actor and Critic parameters \(w_n^0\).
\FOR{each global round \(t = 1, 2, \dots, T\)}
    \STATE \textbf{Local Actor-Critic Training:}
    \FOR{each agent \(n\) in parallel}
        \STATE Run A2C locally for \(\cal T\) timesteps using \(w_n^{t-1}\).
        \STATE Obtain updated local parameters \(w_n^t\).
        \STATE Upload \(w_n^t\) to the central FL server.
    \ENDFOR
    
    \STATE \textbf{Clustering Step:}
    \STATE The server groups agents into \(K\) clusters \( \{C_{(1)},\dots,C_{(K)}\}\) based on similarity of parameters or other relevant attributes.

    \STATE \textbf{Within-Cluster Aggregation:}
    \FOR{each cluster \(C_{(k)}\) in \(\{C_{(1)},\dots,C_{(K)}\}\)}
        \STATE Update $\overline{w}_{(k)}$ based on Equ.~\eqref{equ:clusterlight_avg}
    \ENDFOR
    
    \STATE \textbf{Broadcast to Clusters:}
    \FOR{each agent \(n\)}
        \STATE Determine the cluster \(k\) such that \(n \in C_{(k)}\).
        \STATE Broadcast the weights to the agents according to Equ.~\eqref{equ:clusterlight_broadcast}.
    \ENDFOR
\ENDFOR
\end{algorithmic}
\end{algorithm}

\section{EXPERIMENTS}
\label{sec:exp}
Experiments are conducted on a local workstation equipped with 14 Intel Xeon Processor E5-2690 v4 CPUs and an NVIDIA 2080 GPU with 8 GB memory in Ubuntu 20.04. The learning rate is $0.001$, the batch size is $3000$, the discount factor $\gamma$ is $0.95$, and the length of the roll-out fragment is $240$.

\subsection{Experiment Set-up}
\label{exp-setup}
\begin{figure}[H]
    \centering
    
    \begin{minipage}[c]{0.3\textwidth}
        \centering
        \includegraphics[width=\textwidth]{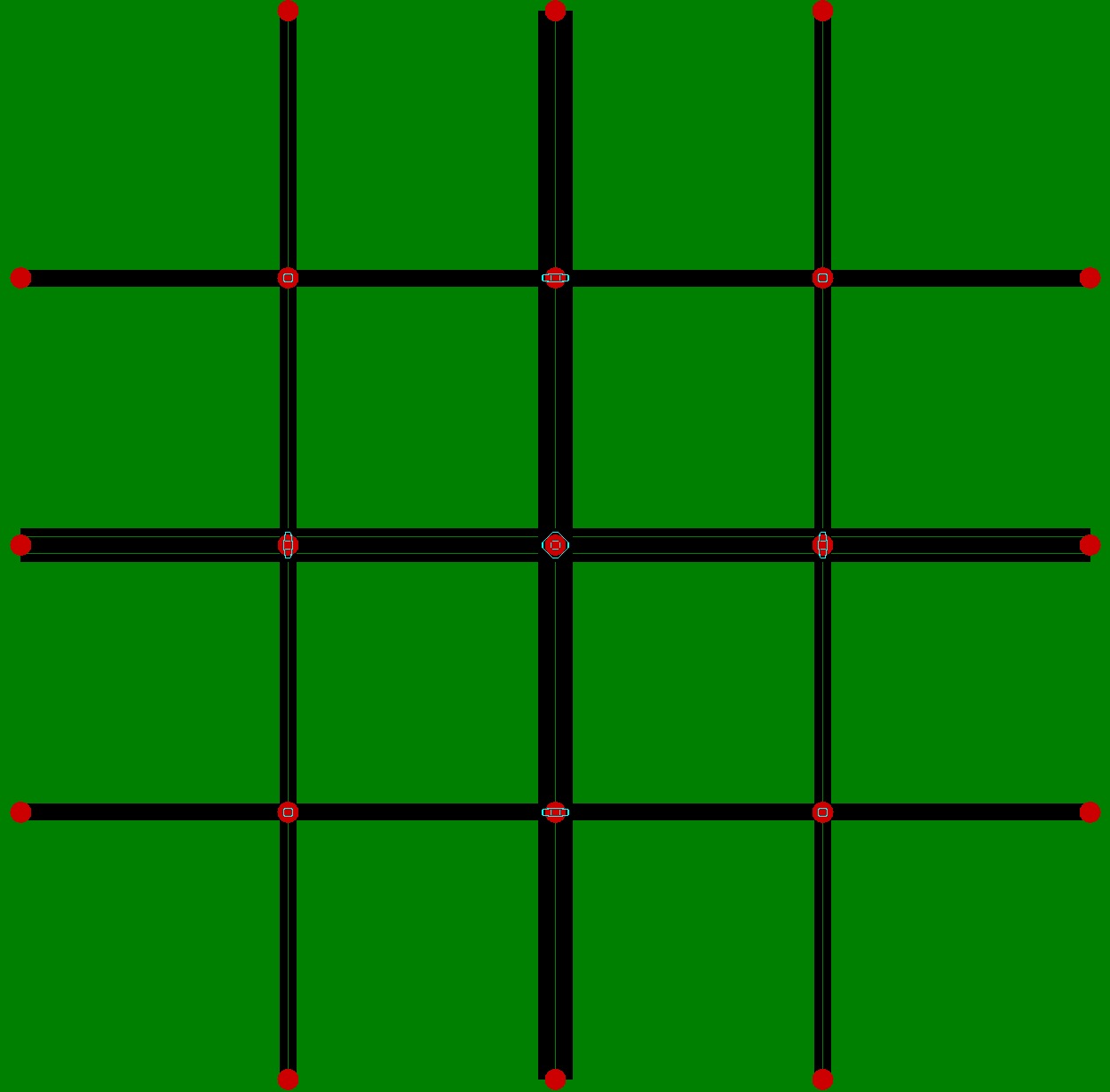}
        \subcaption{3×3 Road Network Layout}
        \label{fig:3x3}
    \end{minipage}
    \hspace{0.01\textwidth}
    \begin{minipage}[c]{0.3\textwidth}
        \centering
        \includegraphics[width=\textwidth]{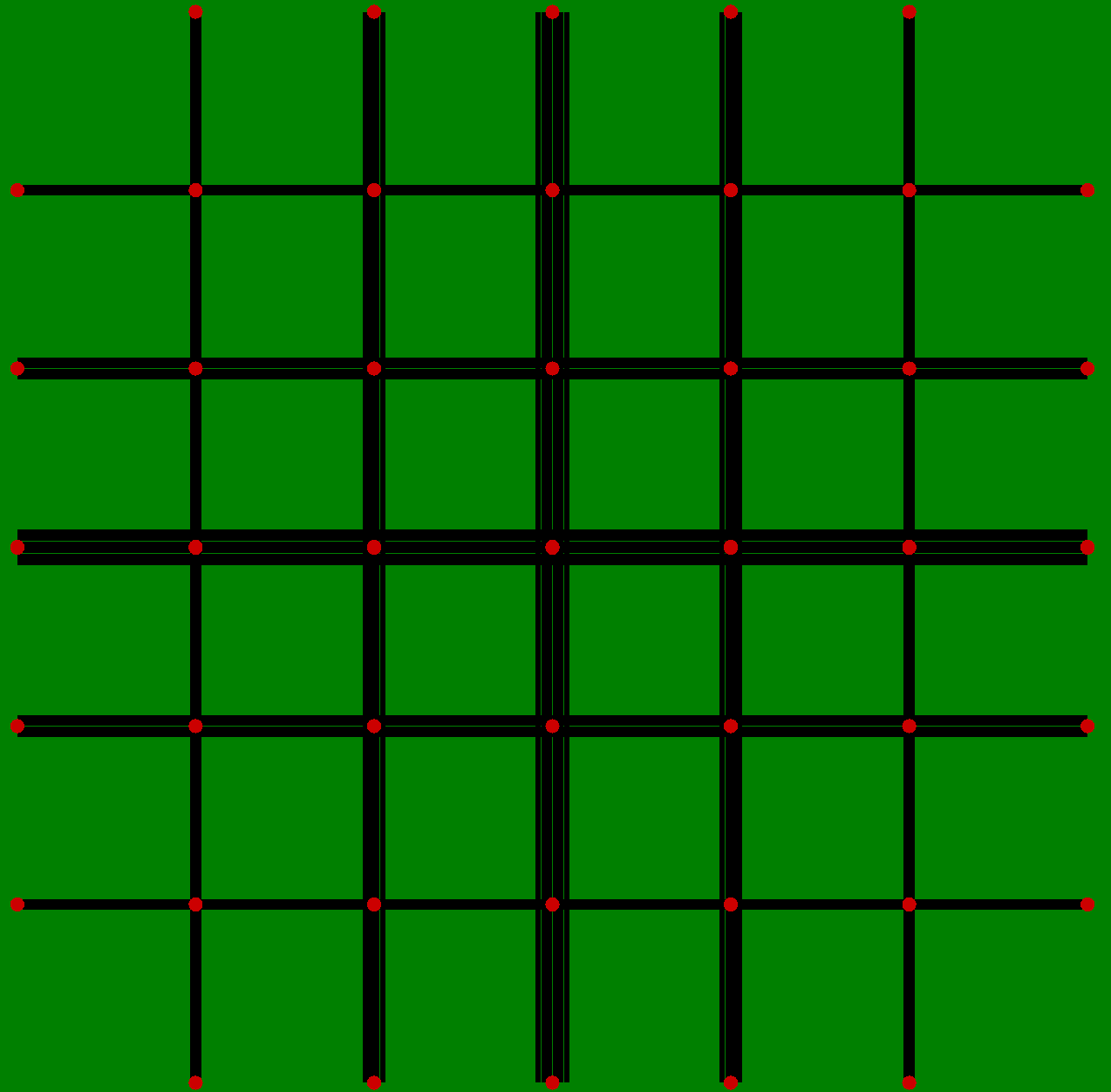}
        \subcaption{5×5 Road Network Layout}
        \label{fig:5x5}
    \end{minipage}
    \begin{minipage}[c]{0.30\textwidth}
        \centering
        \includegraphics[width=\textwidth]{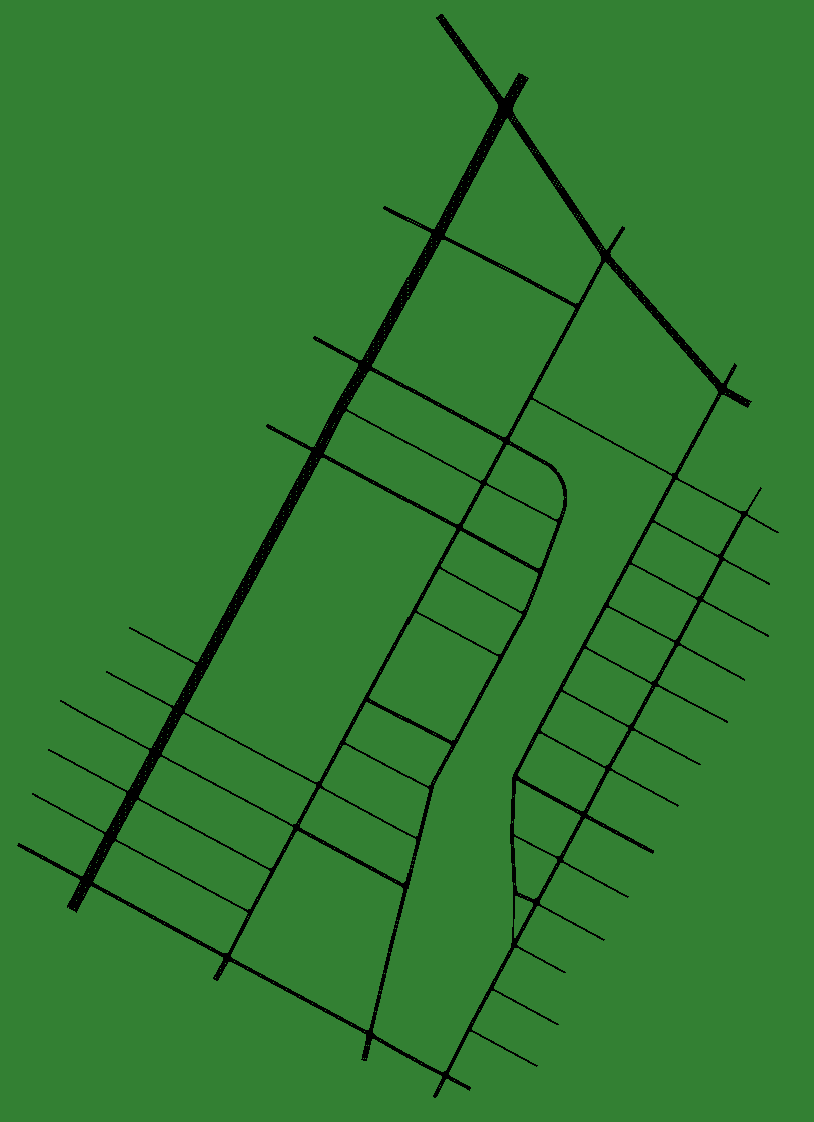}
        \subcaption{Real-world Road Network Layout.}
        \label{fig:real-world}
    \end{minipage}
    \hspace{0.01\textwidth}
    \caption{Illustrations of the synthetic road networks}
    \label{fig:synthetic_networks}
\end{figure}

\begin{figure}[H]
    \centering
    \begin{minipage}[c]{0.35\textwidth}
        \centering
        \includegraphics[width=\textwidth]{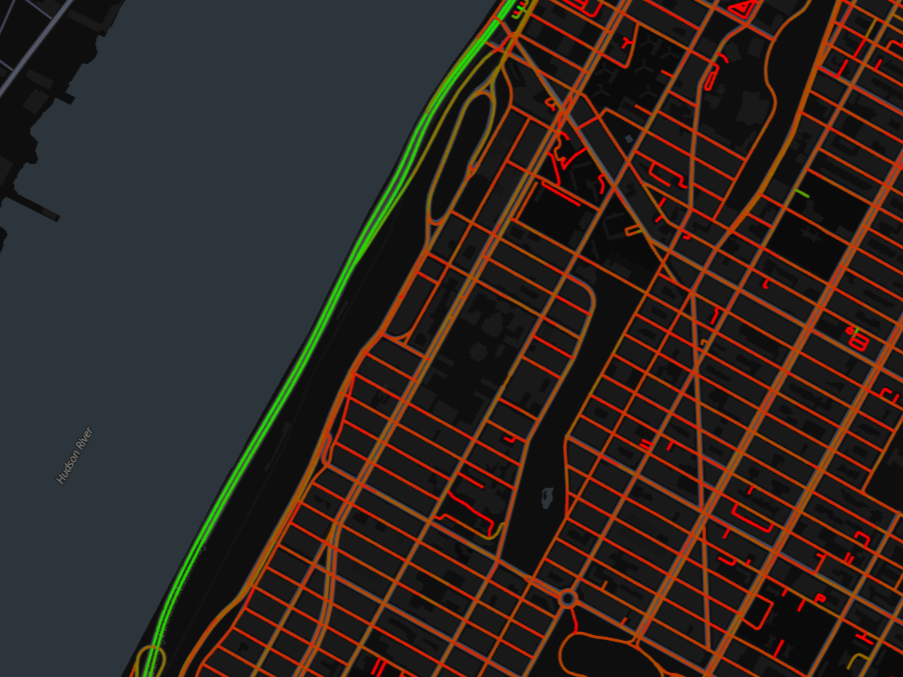}
        \subcaption{Traffic network visualization of the area surrounding Columbia University in NYC}
    \end{minipage}
    \begin{minipage}[c]{0.35\textwidth}
        \centering
        \includegraphics[width=\textwidth]{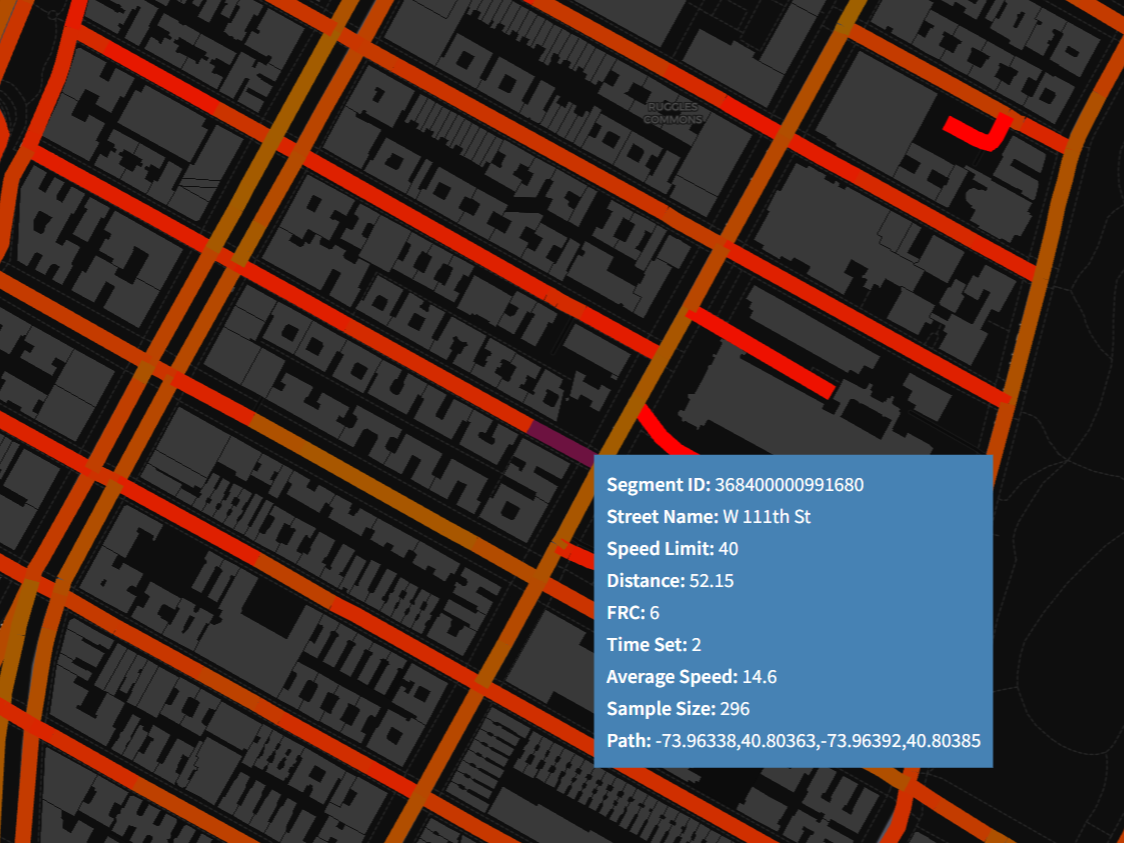}
        \subcaption{Detailed traffic attributes for a specific road segment near Columbia University}
    \end{minipage}
    \caption{(a) A map of the New York City area near Columbia University, generated from the speed info and coordinates from the TomTom dataset.
    (b) A detailed segment of an intersection near Columbia University, highlighting specific traffic attributes. 
    The blue box in (b) provides details such as the segment ID, street name, speed limit, distance, functional road class (FRC), time set, average speed, sample size, and geospatial coordinates of the path. 
    The "sample size" refers to the total number of vehicles recorded passing through a specific road segment during a predefined one-hour time period, aggregated across multiple days over a month.}
    \label{fig:tomtom_cu}
\end{figure}

We utilize Ray's RLlib to train the proposed HFRL methods and evaluate the final policies via simulation. The road networks are shown in Fig.~\ref{fig:synthetic_networks}: (1) Grid 3×3 (see Fig.~\ref{fig:3x3}) (2) Grid 5×5 (see Fig.~\ref{fig:5x5}); (3) Real-world (see Fig.~\ref{fig:real-world}). For the 3×3 network, the primary east-west and north-south routes are designed as two-way, four-lane roads, while all other roads are two-way, two-lane roads; for the 5×5 network, the most central north-south and east-west routes are designed as two-way, six-lane roads, adjacent roads are two-way, four-lane roads, and the outermost roads are two-way, two-lane roads. The real-world network is the road networks near Columbia University generated from the OpenStreetMap\footnote{\url{https://www.openstreetmap.org}, © OpenStreetMap.}. For the Grid 3×3 and Grid 5×5 configurations, we designated the rightmost lane at each intersection as a mixed lane for straight-through and right-turn movements, allowing some vehicles to proceed through the intersection or turn right. The straight-through and right-turn ratios for these lanes are set at 90\% and 10\%, respectively. Regarding the traffic demand, the light demand is $200$ vehicles per lane per hour, while the heavy demand is $400$ vehicles per lane per hour. To build a more realistic simulation on the real-world network, we utilize two data sources to calibrate the simulation: TomTom data from the TomTom MOVE web portal \footnote{\url{https://move.tomtom.com}, © TomTom API.} and manually recorded traffic counts at intersections. We then used data from the TomTom dataset to define input flows at each edge intersection, as illustrated in Fig.~\ref{fig:tomtom_cu}. In addition, we collected real-world traffic data from several intersections over a one-hour period, and these data will also be incorporated into our traffic flow generation file.

We employ the A2C algorithm as the underlying reinforcement learning framework for our proposed HFRL method, with the detailed hyperparameter settings summarized in Tab.~\ref{tab:a2c_config}. A2C is selected because it offers a good balance between training stability and computational efficiency. As a synchronous version of the actor–critic framework, it updates policy and value networks simultaneously, allowing the agents to learn both optimal actions and accurate value estimates. This parallel structure reduces variance in policy gradients compared with purely policy-based methods and enables training stabilization, which is particularly beneficial in multi-agent traffic signal control environments.

\captionof{table}{Simulation Configuration for HFRL Experiments}
\label{tab:simulation_config}
\begin{tabular}{l l}
\hline
\textbf{Parameter} & \textbf{Value / Description} \\
\hline
Simulation software & SUMO v1.19.0 \\
Episode duration & 1000 time steps \\
Total training episodes & 100 (Synthetic networks), 50 (Real-world network) \\
Road networks & Grid 3$\times$3, Grid 5$\times$5, Real-world \\
Lane configurations & See Fig.~\ref{fig:synthetic_networks}, lane length = 100 m \\
Mixed lane ratio (Grid) & 90\% straight, 10\% right-turn \\
Turning ratios (Real-world) & $\approx$ 80\% straight, 15\% right-turn, 5\% left-turn \\
Traffic demand (light) & 200 veh/lane/hour \\
Traffic demand (heavy) & 400 veh/lane/hour \\
Phase structure & 4-phase plan: two opposing green phases with yellow transitions \\
Green phase duration & 42 s (per direction, subject to RL control) \\
Yellow phase duration & 3 s \\
Minimum green time & 4 s \\
Maximum green time & 120 s \\
Cluster number $K$ & 4 \\
Calibration data & TomTom MOVE + manual traffic counts \\
\hline
\end{tabular}

\captionof{table}{A2C Algorithm Hyperparameters}
\label{tab:a2c_config}
\begin{tabular}{l l}
\hline
\textbf{Hyperparameter} & \textbf{Value} \\
\hline
Framework & RLlib (Ray) \\
Policy network architecture & 2 fully connected layers, 256 units each, ReLU activation \\
Optimizer & Adam \\
Learning rate & $1\times10^{-4}$ \\
Discount factor ($\gamma$) & 0.99 \\
Entropy coefficient & 0.01 \\
Value function coefficient & 0.5 \\
Rollout length & 20 steps \\
Batch size & 4,000 \\
Gradient clipping & 40.0 \\
Number of workers & 8 \\
Random seed & 42 \\
\hline
\end{tabular}



\subsubsection{Baseline}

\begin{enumerate}
    \item \textbf{Centralized RL.} 
    In the centralized RL approach, a single policy network serves as a centralized controller that manages all intersections simultaneously. Each intersection periodically sends its local traffic observations to the central controller, which then processes the combined information from the entire network to determine coordinated signal actions for all intersections. Because the controller learns from global observations, it can achieve faster stabilization and more effective coordination across the network. However, this approach requires continuous communication between all intersections and the central controller, leading to significant communication overhead and high computational demand in large-scale systems.

    \item \textbf{Decentralized RL.} 
    Here, each intersection has its own independent RL agent that updates its parameters using only local observations. Although this reduces communication and scales better, lack of direct coordination can lead to suboptimal solutions in some scenarios.

    \item \textbf{FedAvg.} 
    In this federated learning method, agents at each intersection train local models using only their respective observations. After a fixed number of local training steps, these models are uploaded to a central FL server, which aggregates them by taking a weighted average of the parameters. Each local agent then downloads the aggregated (global) model and continues training on its local data, thereby iterating between local updates and global aggregation.
\end{enumerate}

\subsubsection{Performance Metrics}
\label{sec:performance-metric}

\begin{enumerate}

    \item \textbf{Travel Time.}  
The travel time of vehicle $i$ is defined as
\[
T_i = t_i^{\text{arrival}} - t_i^{\text{departure}},
\]
where $t_i^{\text{departure}}$ and $t_i^{\text{arrival}}$ denote the departure and arrival times of vehicle $i$, respectively. The average travel time across all vehicles is then computed as
\[
\overline{T} = \frac{1}{N_v} \sum_{i=1}^{N_v} T_i,
\]
where $N_v$ is the total number of vehicles. Note that in SUMO, the departure time corresponds to the time when a vehicle actually enters the network. Therefore, the computed travel time inherently includes departure delay caused by queueing, spillback, or upstream congestion, which is particularly relevant under high-demand conditions.

\item \textbf{Waiting Time.}  
The waiting time of vehicle $i$ is defined as the total time steps during which its speed $v_i(t)$ is below a small threshold, i.e.,
\[
W_i = \sum_t \mathbb{I}\left[v_i(t) \leq 0.1\,\text{m/s}\right],
\]
where $\mathbb{I}[\cdot]$ is the indicator function. The average waiting time across all vehicles is given by
\[
\overline{W} = \frac{1}{N_v} \sum_{i=1}^{N_v} W_i.
\]
Lower values of $\overline{T}$ and $\overline{W}$ indicate more efficient and coordinated traffic signal control.

    \item \textbf{Communication Cost.}
    This is an estimated communication cost \citep{hudson2022smart} including: (1) transmitting critic and actor parameters from edge devices to the central server; (2) distributing these parameters from the central server back to the edge devices; (3) sending actions from edge devices to traffic lights; (4) relaying observations from traffic lights to edge devices; and (5) transmitting vehicle data from vehicles to the traffic lights.
\end{enumerate}

\noindent
Together, travel time and waiting time indicate the efficiency of the training strategy for traffic signal control, while communication cost quantifies the overhead required to coordinate agents in the system.

\subsection{Synthetic Examples}
We apply the proposed algorithms to synthetic networks, as shown in Fig.~\ref{fig:3x3} and Fig.~\ref{fig:5x5}. The results are analyzed in terms of travel time, waiting time, and clustering patterns.
\subsubsection{Reward Stabilization}
\begin{figure}[H]
        \centering
	\begin{minipage}[c]{0.46\textwidth}
		\centering
		\includegraphics[width=\textwidth]{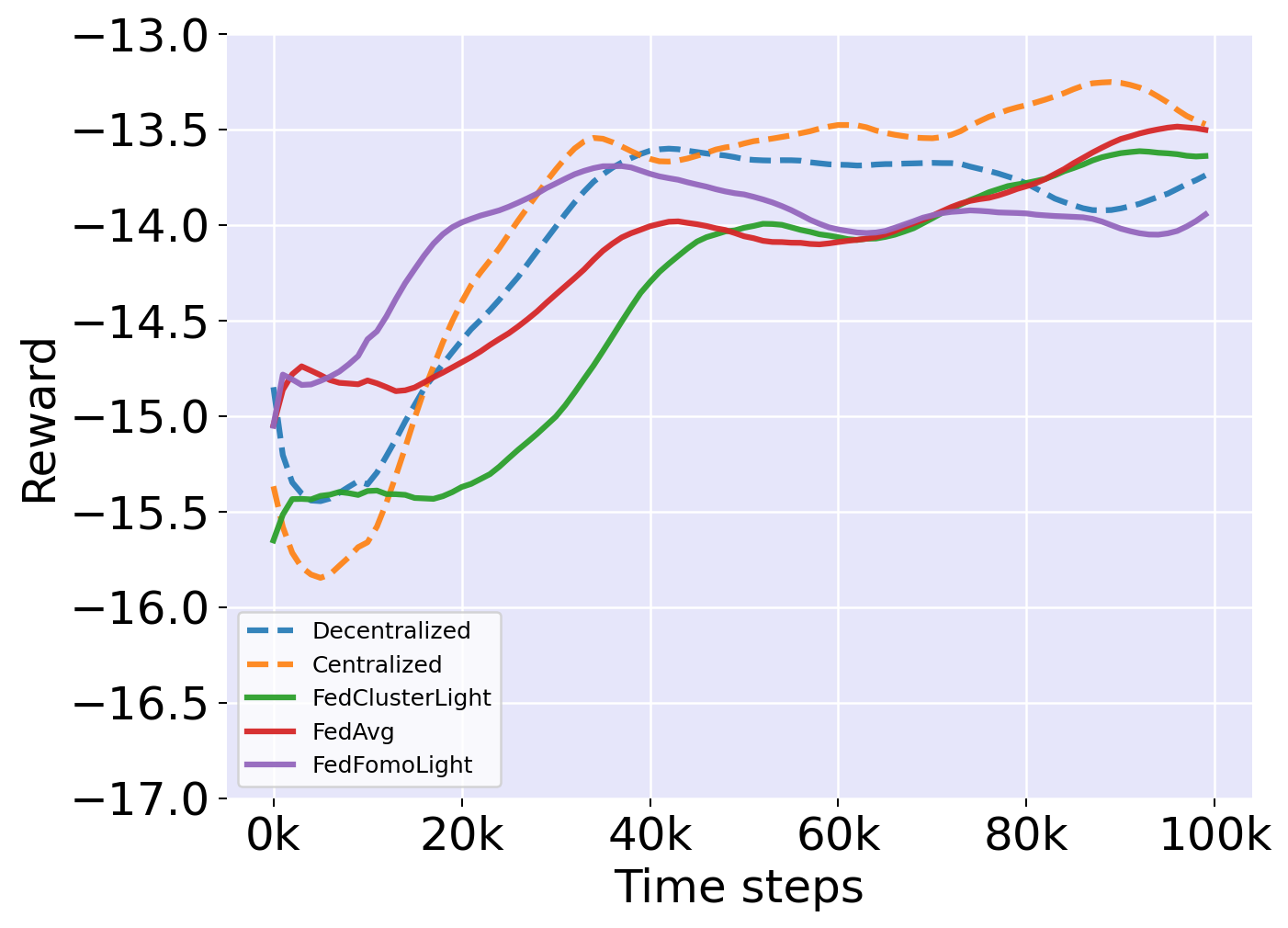}
		\subcaption{Reward curve on Grid 3×3 under light demand}
	\end{minipage} 
	\centering
	\begin{minipage}[c]{0.45\textwidth}
		\centering
		\includegraphics[width=\textwidth]{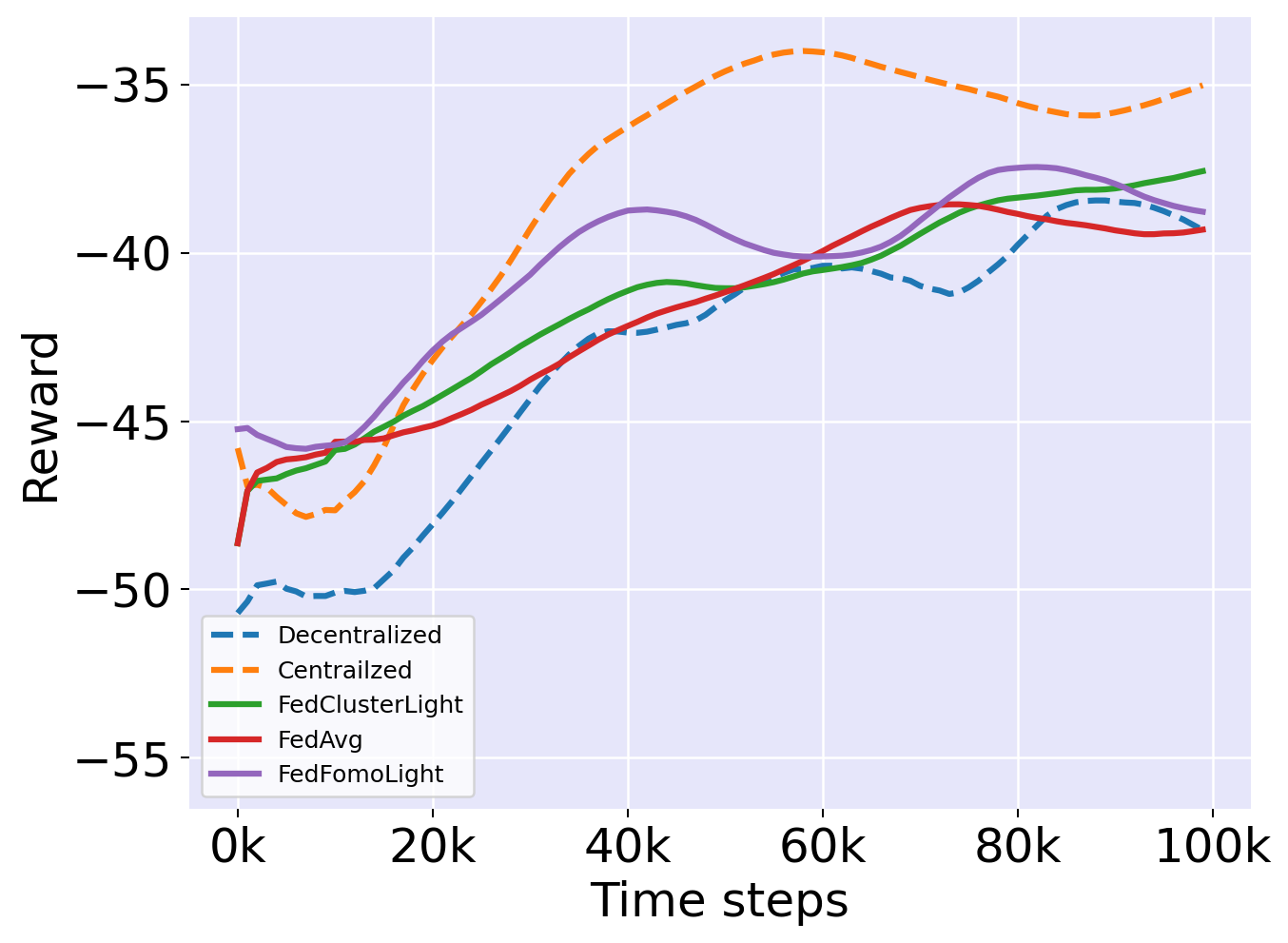}
		\subcaption{Reward curve on Grid 3×3 under heavy demand}
	\end{minipage} 
        \caption{Reward curve on Grid 3×3}
	\label{fig:Test_Reward3x3}
\end{figure}

\begin{figure}[H]
        \centering
	\begin{minipage}[c]{0.45\textwidth}
		\centering
		\includegraphics[width=\textwidth]{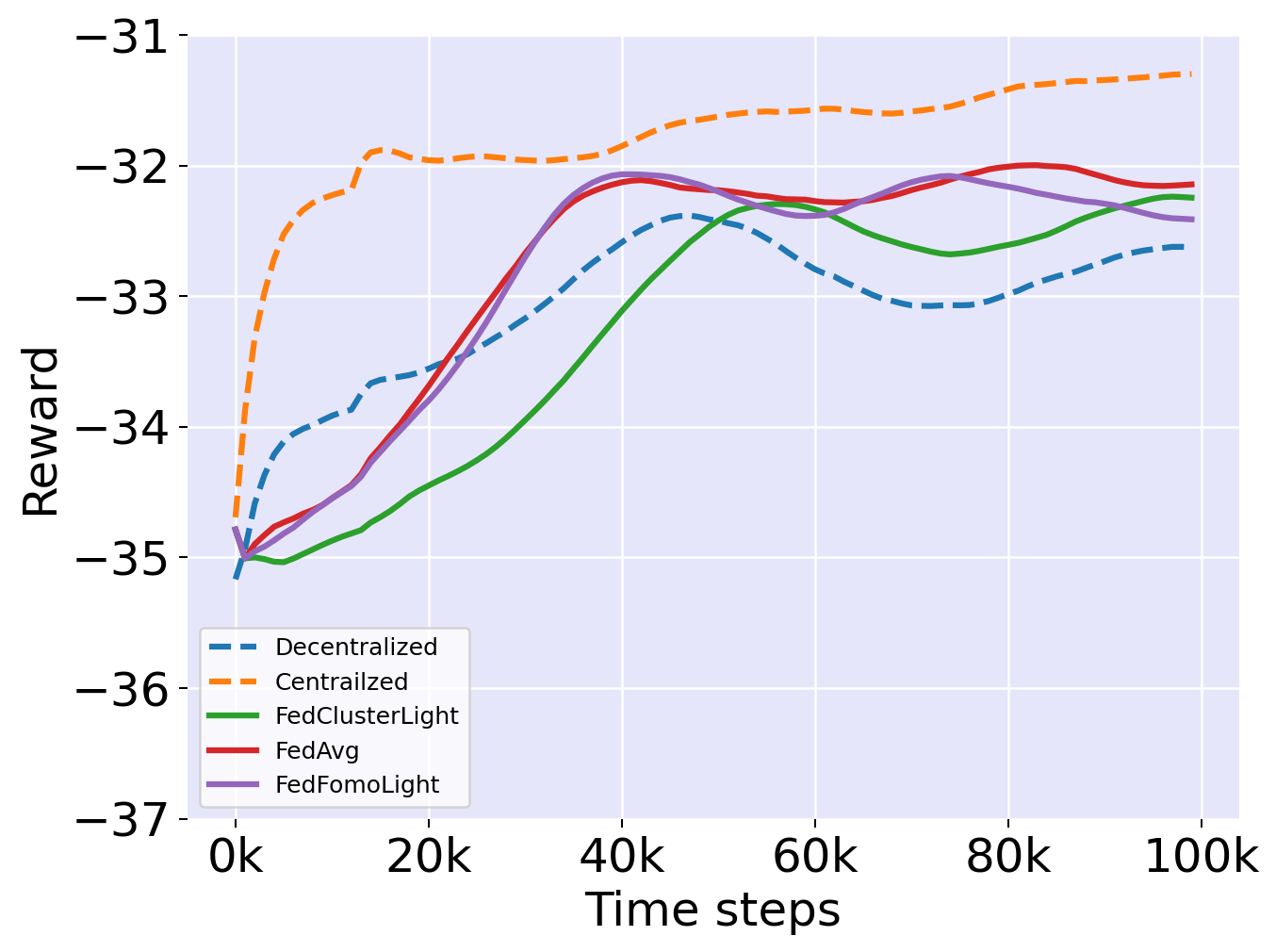}
		\subcaption{Reward curve on Grid 5×5 under light demand}
	\end{minipage} 
	\centering
	\begin{minipage}[c]{0.45\textwidth}
		\centering
		\includegraphics[width=\textwidth]{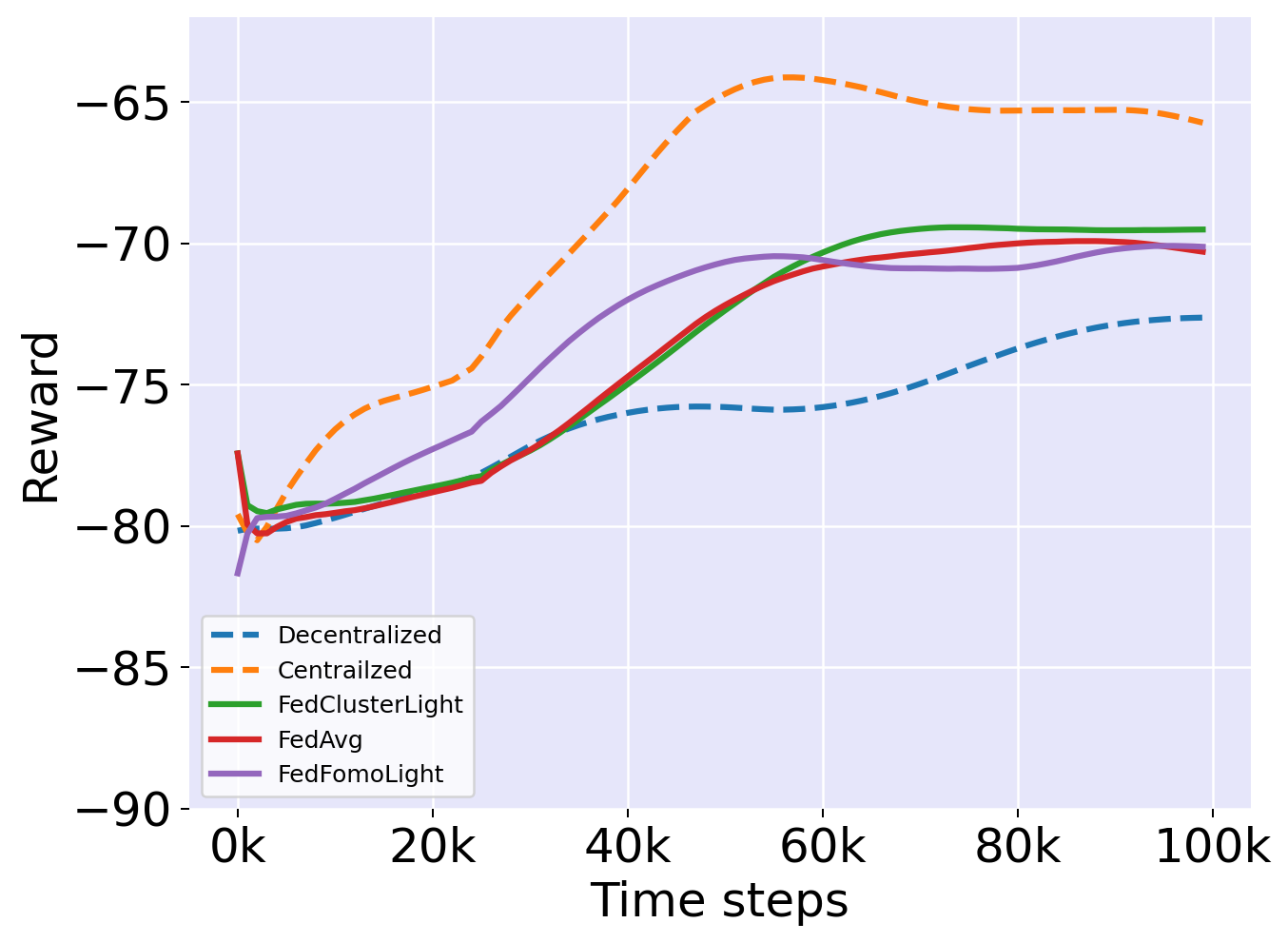}
		\subcaption{Reward curve on Grid 5×5 under heavy demand}
	\end{minipage} 
        \caption{Reward curve on Grid 5×5}
	\label{fig:Test_Reward5x5}
\end{figure}

In Fig. \ref{fig:Test_Reward3x3} and \ref{fig:Test_Reward5x5}, we compare three distinct training strategies across the Grid 3×3 synthetic network under different traffic demands. The figure demonstrates that the centralized RL generally achieves the highest rewards. This superior performance is attributed to its comprehensive access to the information from each agent. However, our focus will be on comparing performance metrics.

\subsubsection{Performance Comparison}

For the Grid 3×3 network, we present comparisons of travel time and waiting time among the proposed FedClusterLight, FedFomoLight, and baseline methods in Tab.~\ref{tab:travel_waiting_time}. The results show that for the Grid 3×3 network under relatively low traffic demand, the centralized method outperforms both the decentralized and FedAvg methods. FedClusterLight achieves results close to those of FedAvg, while the proposed FedFomoLight method ranks second only to the centralized approach. Overall, the centralized method achieves the best performance, surpassing all other baselines. Regarding waiting time, the general trend is consistent with that of travel time.

For the Grid 5×5 network, however, the trend differs. Under light traffic conditions, as shown in Tab.~\ref{tab:travel_waiting_time}, FedFomoLight achieves the best performance, while FedClusterLight also outperforms the other methods. In heavy traffic conditions, the centralized method performs better than the proposed methods, benefiting from complete global observations and relatively low computational and communication overhead, which allows it to coordinate all intersections more effectively.

\begin{table}[htbp]

    \centering
    \caption{Travel Time and Waiting Time for Different Algorithms on Various Networks}
    \label{tab:travel_waiting_time}
    \setlength{\tabcolsep}{6pt}
    \begin{tabular}{llcccccc}
        \hline
        \multirow{2}{*}{\textbf{Algorithm}} & \multirow{2}{*}{\textbf{Volume}} 
        & \multicolumn{2}{c}{\textbf{3$\times$3 Network}} 
        & \multicolumn{2}{c}{\textbf{5$\times$5 Network}} 
        & \multicolumn{2}{c}{\textbf{Real-world Network}} \\
        \cline{3-8}
        & & Travel~$\downarrow$ & Waiting~$\downarrow$ & Travel~$\downarrow$ & Waiting~$\downarrow$ & Travel~$\downarrow$ & Waiting~$\downarrow$ \\
        \hline
        \multirow{2}{*}{FedFomoLight}    
            & Light & 53.53 & 10.87 & $\boldsymbol{82.23}$ & $\boldsymbol{17.04}$ & $\boldsymbol{144.73}$ & $\boldsymbol{34.20}$ \\
            & Heavy & 71.58    & 21.15    & 97.81    & 26.84    & /     & /    \\
        \multirow{2}{*}{FedClusterLight} 
            & Light & 57.90 & 12.85 & 83.44 & 17.81 & 148.70 & 36.62 \\
            & Heavy & 69.65    & 19.6    & 100.22    & 30.46    & /     & /    \\
        \multirow{2}{*}{FedAvg}          
            & Light & 58.57 & 13.42 & 85.77 & 18.36 & 157.41 & 37.16 \\
            & Heavy & 77.85    & 23.35    & 105.61    & 31.04    & /     & /    \\
        \multirow{2}{*}{Decentralized}   
            & Light & 57.18 & 12.46 & 87.23 & 23.40 & 643.75 & 501.38 \\
            & Heavy & 78.31    & 28.25    & 112.11    & 66.45   & /     & /    \\
        \multirow{2}{*}{Centralized}     
            & Light & $\boldsymbol{50.11}$ & $\boldsymbol{8.03}$  & 83.20 & 17.62 & 356.01 & 247.95\\
            & Heavy & $\boldsymbol{69.32}$    & $\boldsymbol{19.53}$    &  $\boldsymbol{96.08}$    & $\boldsymbol{24.85}$   & /     & /    \\
        \multirow{2}{*}{MPlight}         
            & Light & 62.39    & 80.41    & 119.83    & 30.60    & 710.27     & 210.09    \\
            & Heavy & 73.35    & 47.83    & 138.54    & 38.85    & /     & /    \\
        \multirow{2}{*}{Fixed time}      
            & Light & 86.75    & 17.37    & 150.19    & 96.11    & 767.12     & 311.05    \\
            & Heavy & 103.55    & 78.45    & 209.25    & 159.34    & /     & /    \\
        \multirow{2}{*}{Actuated}        
            & Light & 75.24    & 26.09    & 142.32    & 55.02    & 548.35     & 138.29    \\
            & Heavy & 99.45    & 34.57    & 176.83    & 100.98    & /     & /    \\
        \hline
    \end{tabular}
\end{table}

We compare communication costs across different traffic networks under various traffic demand scenarios in Tab.~\ref{tab: Peformance_GRID3X3} and Tab.~\ref{tab: Peformance_GRID5x5}. The centralized method typically requires uploading all local observations to a central server for computation, which leads to significantly higher communication costs. In contrast, FRL and the decentralized method generally maintain similarly low communication costs. FRL methods, due to the allowance for model information exchange, ensure more stable communication costs during training.

\begin{table}[H]
    \centering
    \caption{Communication Cost on Grid 3×3.}
    \label{tab: Peformance_GRID3X3}
    \begin{tabular}{lcc}
        \hline
        & \multicolumn{2}{c}{Grid 3×3 (Bytes)} \\
        \cline{2-3}
        Methods & Light Traffic Demand & Heavy Traffic Demand\\
        \hline
        FedFomoLight & 67337.19±429.06 & 113309.57±2480.82\\
        FedClusterLight & 67660.77±720.60 & 114201.89±2373.63\\
        FedAvg & 67791.21± 693.72 & 114333.80±2594.93\\
        Decentralized & 67665.12±927.51 & 123520.54±3573.32\\
        Centralized & 86504.45±1164.54 & 130151.65±4976.57\\
        \hline
    \end{tabular}
\end{table}


\begin{table}[H]
    \centering
    \caption{Communication Cost on Grid 5×5.}
    \label{tab: Peformance_GRID5x5}
    \begin{tabular}{lcc}
        \hline
        & \multicolumn{2}{c}{Grid 5×5 (Bytes)} \\
        \cline{2-3}
        Methods & Light Traffic Demand & Heavy Traffic Demand\\
        \hline
        FedFomoLight & 191753.69±1617.01 & 279805.56±3316.91\\
        FedClusterLight & 192239.33±1187.04 & 279871.82±4698.67\\
        FedAvg & 191341.04±1198.80 & 279109.07±5050.98\\
        Decentralized & 193707.06±2836.18 & 320436.20±13598.34\\
        Centralized & 243402.81±1226.49 & 326909.85±5983.36\\
        \hline
    \end{tabular}
\end{table}

We also report the average training time per epoch in Tab.~\ref{tab: training_time}. 
As shown in the table, the training time for distributed methods does not differ significantly from that of the centralized approach. 
In fact, the proposed methods achieve slightly lower training times due to their improved performance, which accelerates the stabilization of simulation trials. To further examine scalability, we conducted additional experiments on larger networks with 100 and 150 intersections on FedFomoLight. 
The average training time per epoch increases from approximately 550 s to 689.15 s as the number of intersections grows.
As shown in Fig. 7, the training time increases approximately linearly with the number of intersections. This demonstrates that the proposed HFRL framework can scale to large networks while maintaining reasonable computational efficiency. 
Although communication and synchronization overhead inevitably grow with network size, the hierarchical grouping in FedFomoLight and FedClusterLight effectively mitigates these issues by limiting parameter aggregation within smaller clusters, rather than across all agents globally.

\begin{figure}[H]
  \centering
  \includegraphics[width=0.8\linewidth]{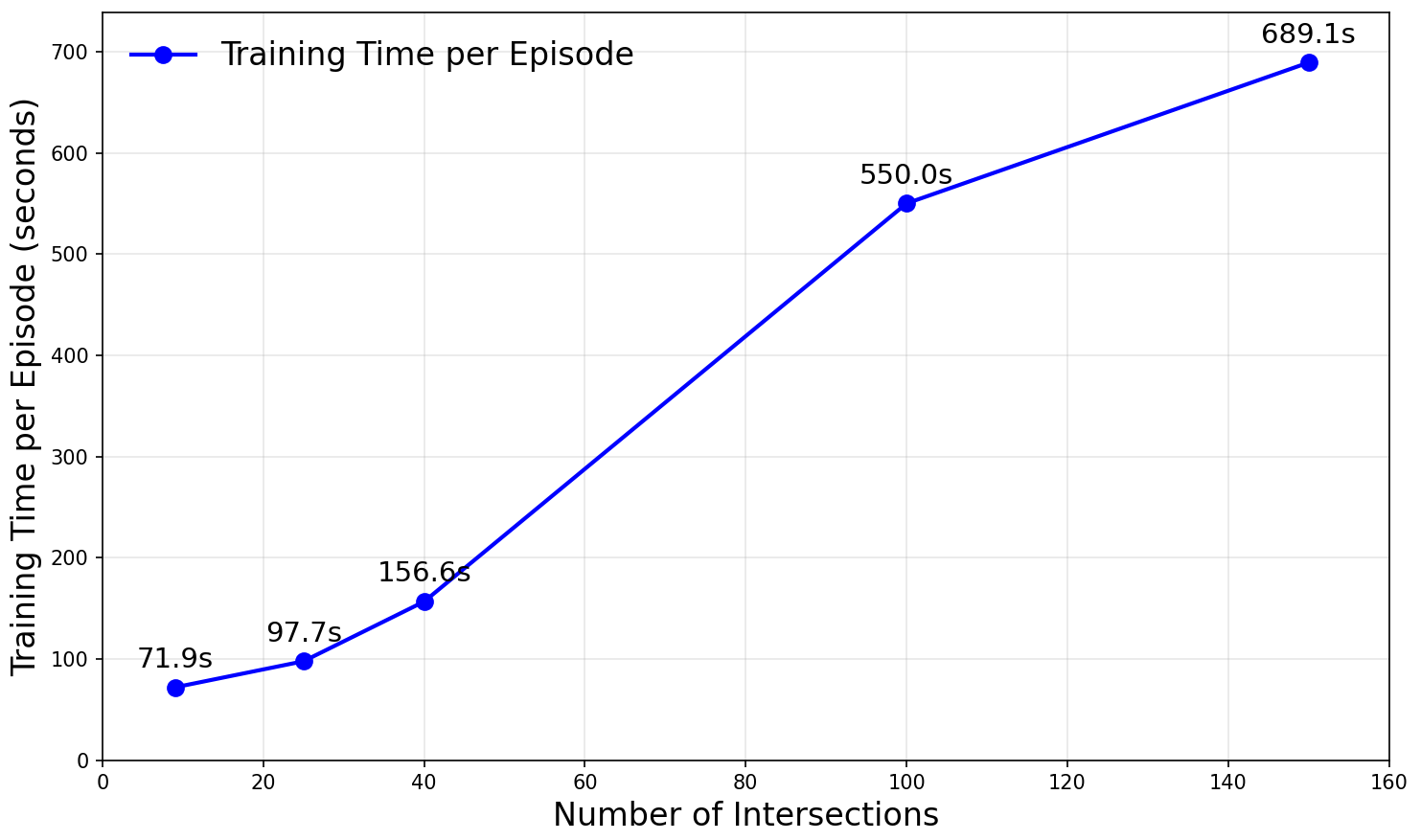}
  \caption{Scalability of Training Time with Increasing Number of Intersections}
  \label{fig:train_epi}
\end{figure}

\begin{table}[H]
    \centering
    \caption{Average Training Time on each Epoch.}
    \label{tab: training_time}
    \begin{tabular}{lccc}
        \hline
        & \multicolumn{3}{c}{Training Time Each Epoch (Seconds)} \\
        \cline{2-4}
        Methods & Grid 3x3 & Grid 5x5 & Cosmos\\
        \hline
        FedFomoLight & 71.8797 & 97.6731 & 156.6420\\
        FedClusterLight & 72.3310 & 99.0668 & 156.6511\\
        FedAvg & 71.3830 & 101.1719 & 159.7703\\
        Decentralized & 74.0686 & 99.8479 & 159.9270\\
        Centralized & 59.6006 & 87.8595 & 145.9587\\
        \hline
    \end{tabular}
\end{table}

\subsubsection{Sensitivity Analysis}

\begin{figure}[H]
        \centering
        \begin{minipage}[c]{0.4\textwidth}
		\centering
		\includegraphics[width=\textwidth]{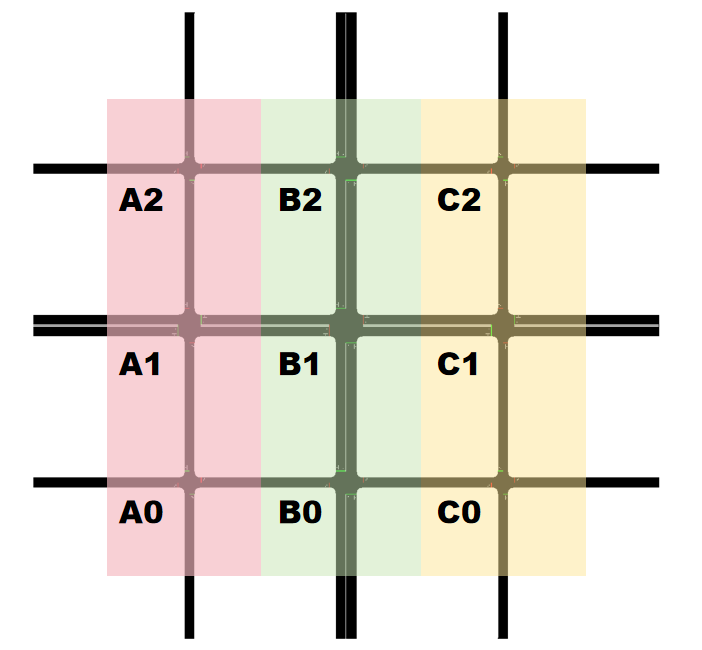}
		\subcaption{Scenario 1 on 3x3 Grid}
	\end{minipage} 
	\begin{minipage}[c]{0.4\textwidth}
		\centering
		\includegraphics[width=\textwidth]{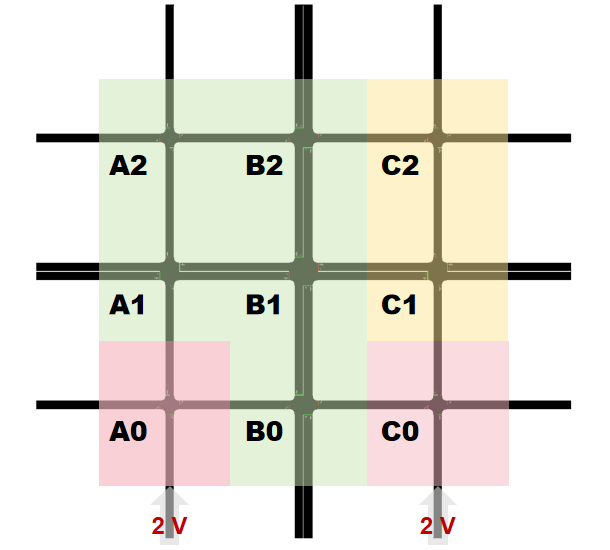}
		\subcaption{Scenario 2 on 3x3 Grid}
	\end{minipage} 
	\centering
	\begin{minipage}[c]{0.4\textwidth}
		\centering
		\includegraphics[width=\textwidth]{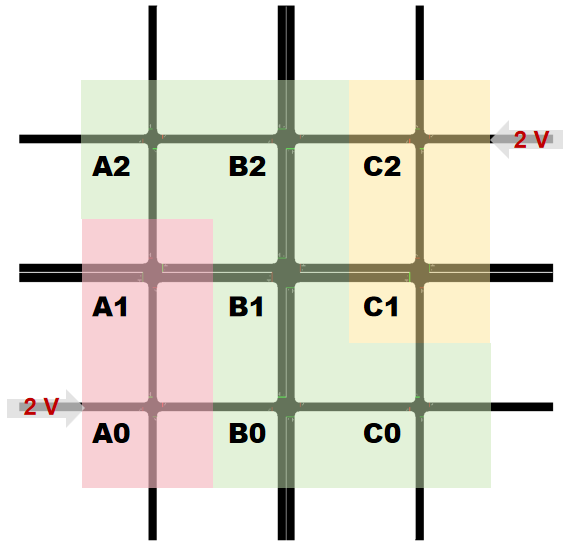}
		\subcaption{Scenario 3 on 3x3 Grid}
	\end{minipage} 
        \begin{minipage}[c]{0.4\textwidth}
		\centering
		\includegraphics[width=\textwidth]{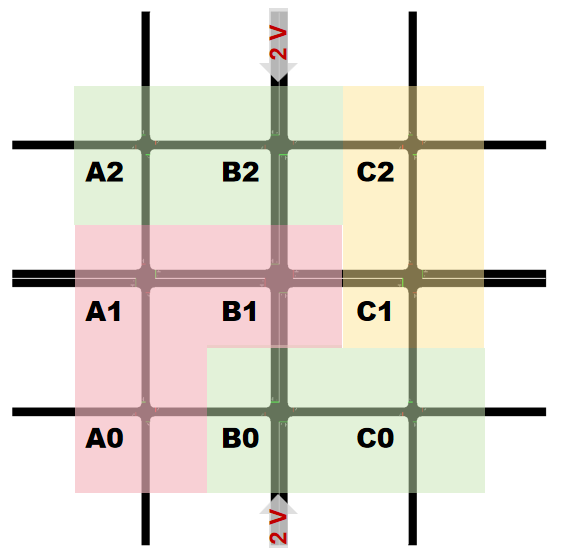}
		\subcaption{Scenario 4 on 3x3 Grid}
	\end{minipage} 
        \caption{Cluster distribution under different traffic demand}
	\label{fig:cluster_sensity}
\end{figure}

This analysis aims to examine the clustering effects of grouping patterns on local models. To this end, we simulated four traffic scenarios by adjusting traffic demand at different intersections and visualized the clustering results from the final round of HFRL. Specifically, this experiment was conducted on a 3×3 network. As illustrated in Fig.~\ref{fig:cluster_sensity}, traffic demand at certain network entrances was doubled, indicated by $2V$, where $V$ is 300 vehicles per lane per hour. After training, we visualized the grouping distribution generated by FedClusterLight.The results clearly show that when traffic demand is doubled on certain roads, the corresponding intersections tend to cluster into the same group. For instance, in Fig.\ref{fig:cluster_sensity}(b), intersections A0 and C0 are grouped together. Similarly, in Fig.\ref{fig:cluster_sensity}(d), intersections B2 and B0 are assigned to the same group. In contrast, in Fig.~\ref{fig:cluster_sensity}(c), intersections C2 and A0, which are significantly distant from each other, are assigned to different groups. The sensitivity analysis explores the impact of varying traffic demands on the clustering behavior of local models within a 3×3 network. By simulating multiple traffic scenarios with adjusted demand levels at specific intersections, the analysis evaluates how changes in traffic demand influence the grouping patterns formed during the final round of training. This helps reveal the relationship between traffic dynamics and model clustering within the framework.

\begin{figure}[H]
    \centering
    \begin{minipage}[c]{0.45\textwidth}
        \centering
        \includegraphics[width=\textwidth]{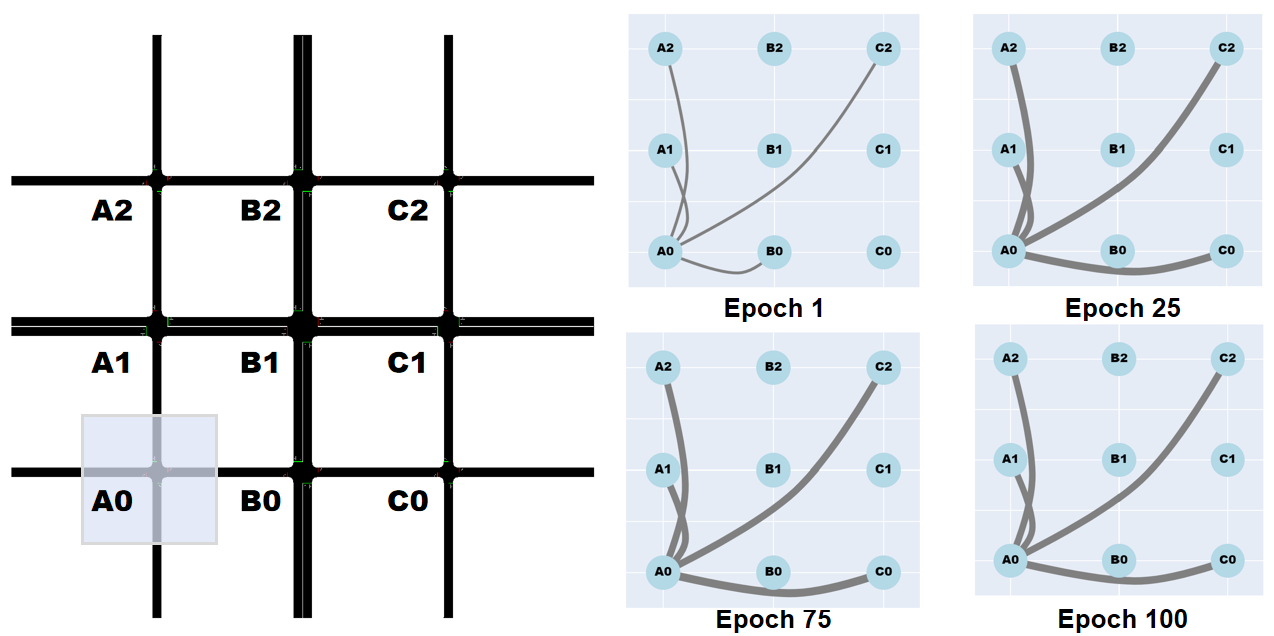}
        \subcaption{Importance evolves over training epochs of A0}
        \label{fig:change_resultA0}
    \end{minipage}\hfill
    \begin{minipage}[c]{0.45\textwidth}
        \centering
        \includegraphics[width=\textwidth]{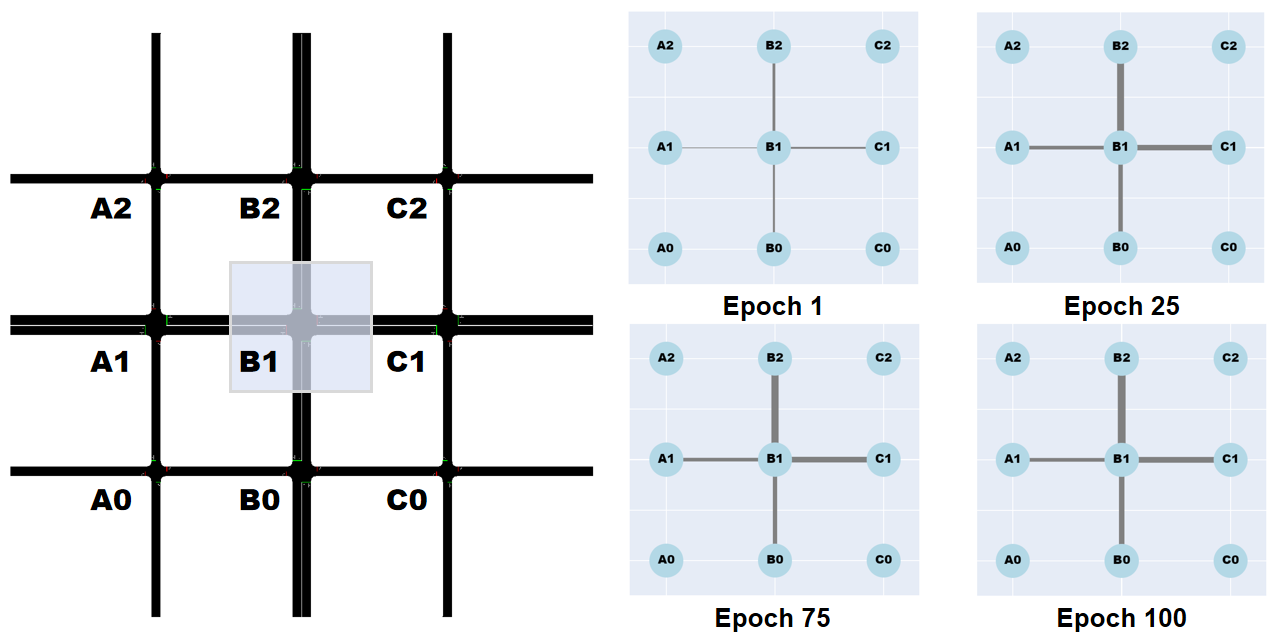}
        \subcaption{Importance evolves over training epochs of B1}
        \label{fig:change_resultB1}
    \end{minipage}

    \vspace{1em}  

    \begin{minipage}[c]{0.45\textwidth}
        \centering
        \includegraphics[width=\textwidth]{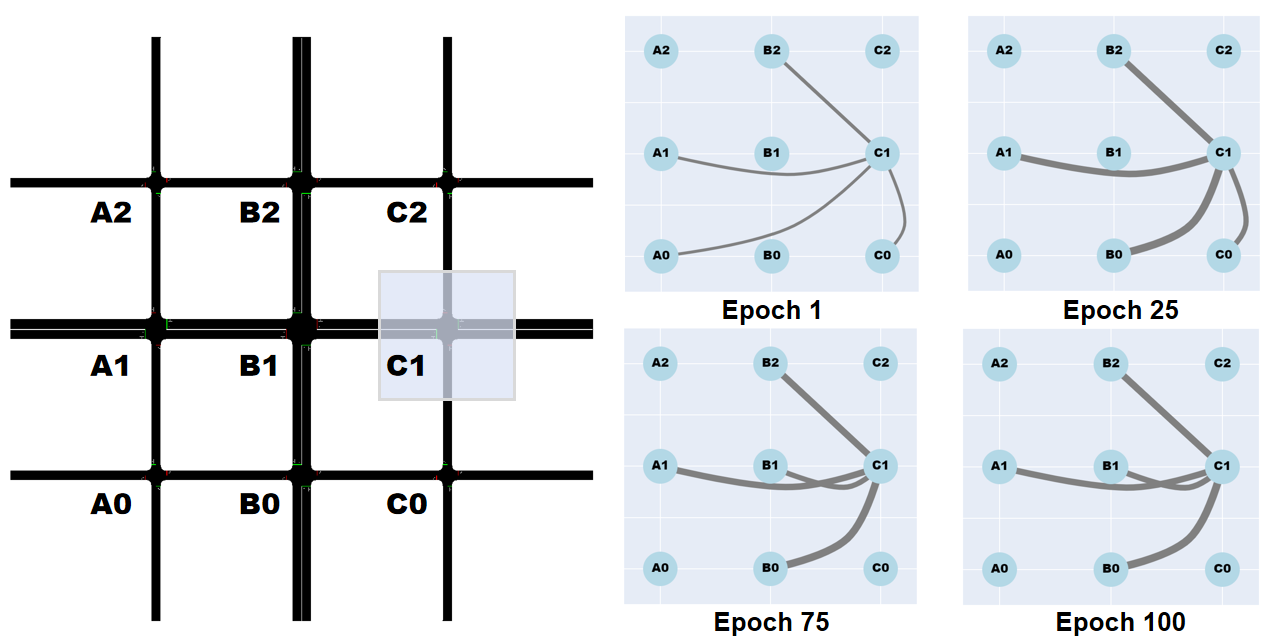}
        \subcaption{Importance evolves over training epochs of C1}
        \label{fig:change_resultC1}
    \end{minipage}

    \caption{Importance evolves over training epochs under different scenarios: (a) A0, (b) B1, and (c) C1}
    \label{fig:combined_weight_changes}
\end{figure}

Fig.~\ref{fig:combined_weight_changes} illustrates how the relative importance of different intersections changes during training. As training progresses, intersections located in adjacent or symmetric locations exhibit increasingly similar neural network parameters. This suggests that the road network topology (i.e., input traffic demand patterns) significantly influences the performance of our proposed method. Using the signal light network parameters of a 3×3 grid as an example, we selected the four most similar intersections for each intersection to illustrate our findings, as shown in Tab.~\ref{tab:net revenue table}. The results indicate that the intersections at the corners exhibit high similarity, while those located along the edges of the main road follow a similar pattern. In addition, the central intersection shows a strong resemblance to the nearest main road edge intersection.






\begin{table}[H]
	\caption{Top 4 similar intersections}\label{tab:net revenue table}
        \begin{center}
            \begin{tabular}{c c}
            \hline
		Intersection ID& Most Similar Intersections (Top 4)\\\hline
            A0& A2, C0, C2, A1  \\
            A1& B0, C1, B2, C0  \\
            A2& A0, C2, C0, A1 \\
            B0& A1, C1, B2, C0 \\
            B1& B2, C1, B0, A1 \\
            B2& B2, C1, B0, A1 \\
            C0& A0, A1, C2, A2 \\
            C1& A1, B0, B2, B1 \\
            C2& A0, A2, C0, A1 \\\hline
	\end{tabular}
        \end{center}
\end{table}

\begin{figure}[H]
  \centering
  \includegraphics[width=0.9\linewidth]{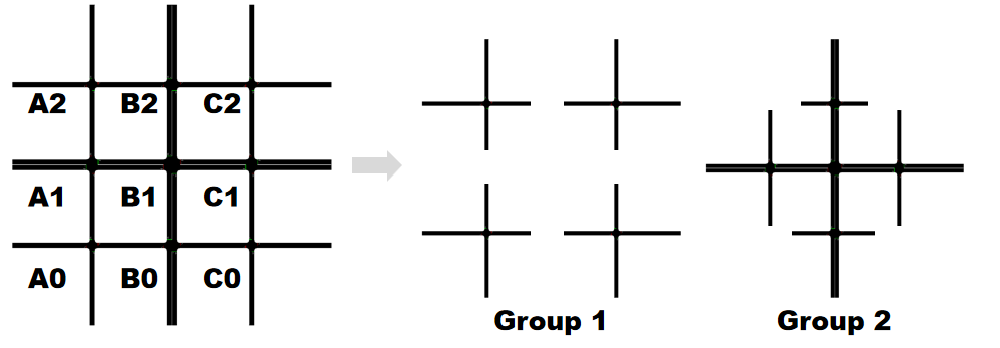}
  \caption{Intersection grouping pattern for Grid 3×3}
  \label{fig:heatmap_result}
\end{figure}

 We calculate similarity based on trained model weights and apply hierarchical clustering to identify two distinct agent groups, as shown in Fig.~\ref{fig:heatmap_result}. We also track the training progression of FedClusterLight by computing local model similarity at each communication round and applying hierarchical clustering to the resulting similarity matrix. To clearly illustrate the evolution of grouping patterns, we select representative frames from key training stages (the 1st, 25th, 75th, and 100th epochs), as depicted in Fig.~\ref{fig:heatmap_result}.

\begin{figure}[H]
  \centering
  \includegraphics[width=0.9\linewidth]{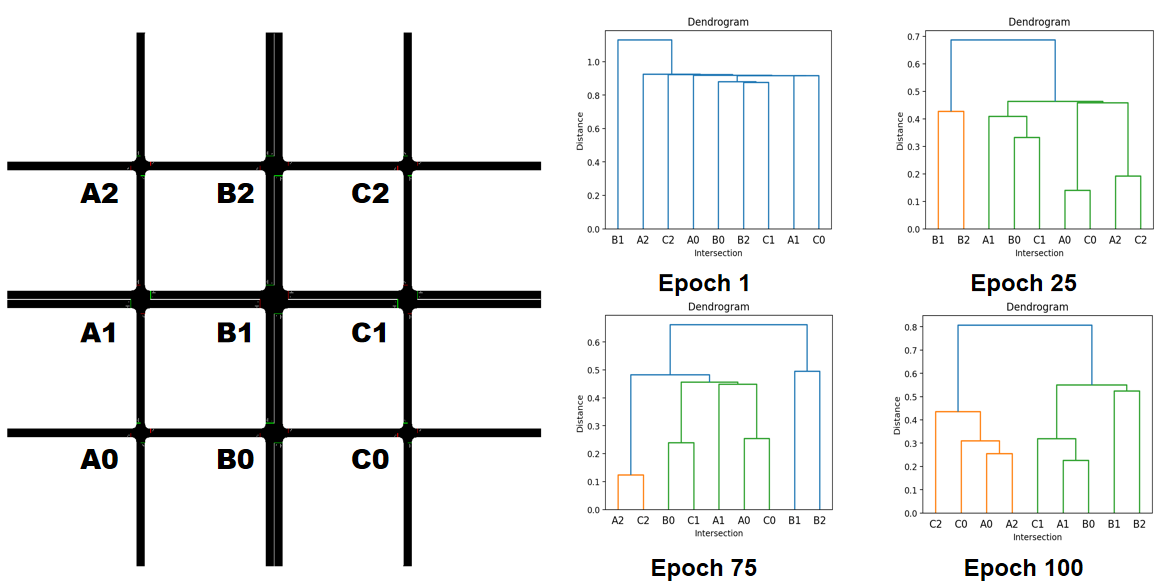}
  \caption{Grouping pattern evolves over training epochs on Grid 3×3}
  \label{fig:HC_weights}
\end{figure}

\subsection{Real-world Experiments}
\begin{figure}[H]
  \centering
  \includegraphics[width=0.33\linewidth]{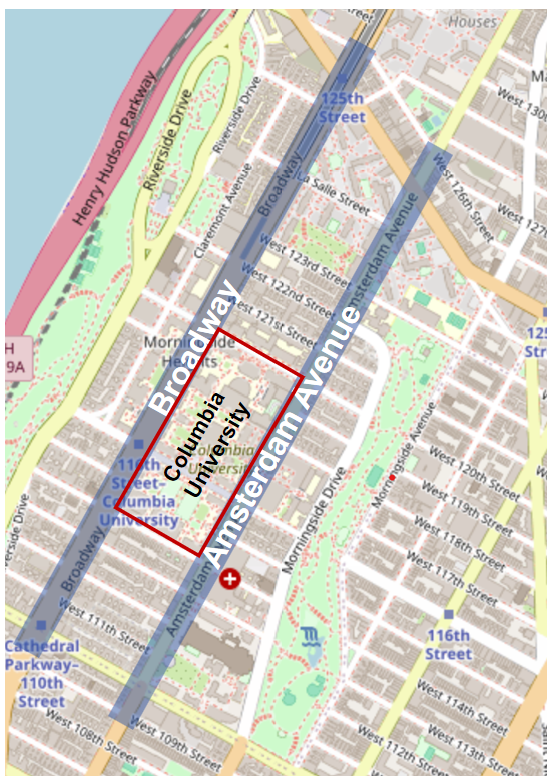}
  \caption{Real-world experiment location}
  \label{fig:real-world-location}
\end{figure}

In addition to the synthetic examples, we also apply the proposed algorithms to real-world networks near Columbia University, as shown in Fig.~\ref{fig:real-world-location}. We configure the traffic demand using the methodology in Section ~\ref{exp-setup} to calibrate the simulation. The results are analyzed in terms of travel time, waiting time, and clustering patterns.

\subsubsection{Reward Stabilization}
\begin{figure}[H]
    \centering
    \includegraphics[width=0.55\linewidth]{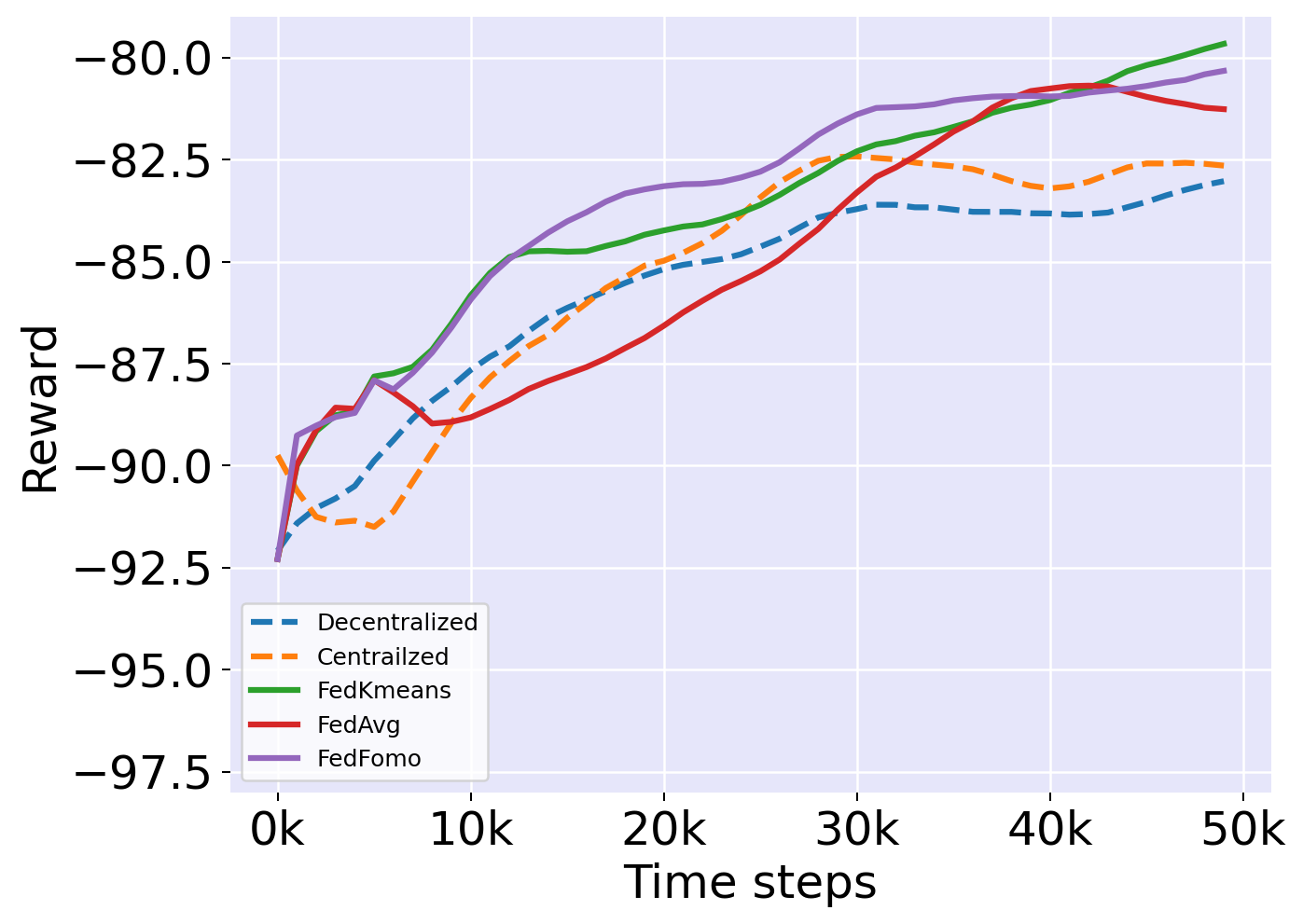}
    \caption{Reward for real-world experiments}
    \label{fig:Test_reward_real}
\end{figure}

Fig.~\ref{fig:Test_reward_real} presents the reward stabilization curves. The training rewards of all algorithms reach a stable plateau within the fixed number of episodes, 
with only minor oscillations near the stabilized stage. Such small fluctuations are common in FRL due to heterogeneous local environments and asynchronous updates~\citep{Qi2021,Xie2022,Emara2024}. The decentralized method has the lowest reward. Our proposed methods outperform other methods in evaluation tests, as discussed in the next section.

\subsubsection{Performance Comparison}

Tab.~\ref{tab:travel_waiting_time} shows the average travel and waiting time in a real-world traffic network, where FedFomoLight achieves the best performance with both metrics. FedClusterLight and FedAvg perform similarly, slightly worse than FedFomoLight, but still significantly better than the other methods. In contrast, the decentralized method performs the worst, exhibiting the highest travel time, while the centralized method also performs poorly. The results indicate that although the proposed methods may not surpass the centralized framework in synthetic networks, they exhibit significantly better performance in real-world networks. This is likely due to the complexity of real-world networks, where the centralized method may not effectively capture the diverse characteristics of intersections within the hierarchical structure.
Tab.~\ref{tab:real-comm} presents the estimated communication cost, calculated using the method in Sec.~\ref{sec:performance-metric}.
The results presented in the Tab.~\ref{tab:Performance-Real} demonstrate that our proposed algorithms, FedClusterLight and FedFomoLight, achieve relatively lower communication costs than the centralized method and FedAvg. Although the decentralized method eliminates the need for communication with a central server, it performs worse than the other methods in real-world settings, as shown in Table 6. As a result, more vehicles remain active on the road, leading to non-negligible communication overhead. This outcome differs from the synthetic experiments, in which the decentralized method exhibits competitive performance and correspondingly lower communication costs.

\begin{table}[H]

	\caption{Communication Cost for Real-world Grid.}\label{tab:Performance-Real}
        \begin{center}
            \begin{tabular}{c c}
            \hline
		 Methods &Real-world Network (Bytes)\\\hline
        FedFomoLight &367212.35±10175.66\\
        FedClusterLight &370573.72±11440.88\\
        FedAvg & 370943.27±9614.38\\
        Decentralized &594852.79±25315.81\\
        Centralized &446646.21±15198.33\\
        \hline
	\end{tabular}
        \end{center}
        \label{tab:real-comm}

\end{table}

\subsubsection{Sensitivity Analysis}
\begin{figure}[H]
  \centering
  \includegraphics[width=1.0\linewidth]{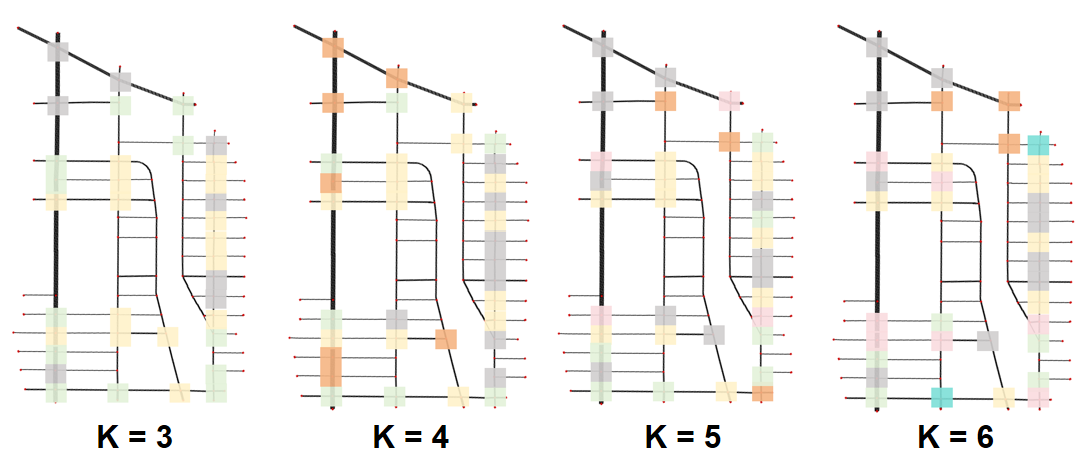}
  \caption{Cluster Patterns of FedClusterLight in the Real-World Network with Different $K$}
  \label{fig:Fedclusterlight}
\end{figure}

\begin{figure}[H]
  \centering
  \includegraphics[width=0.7\linewidth]{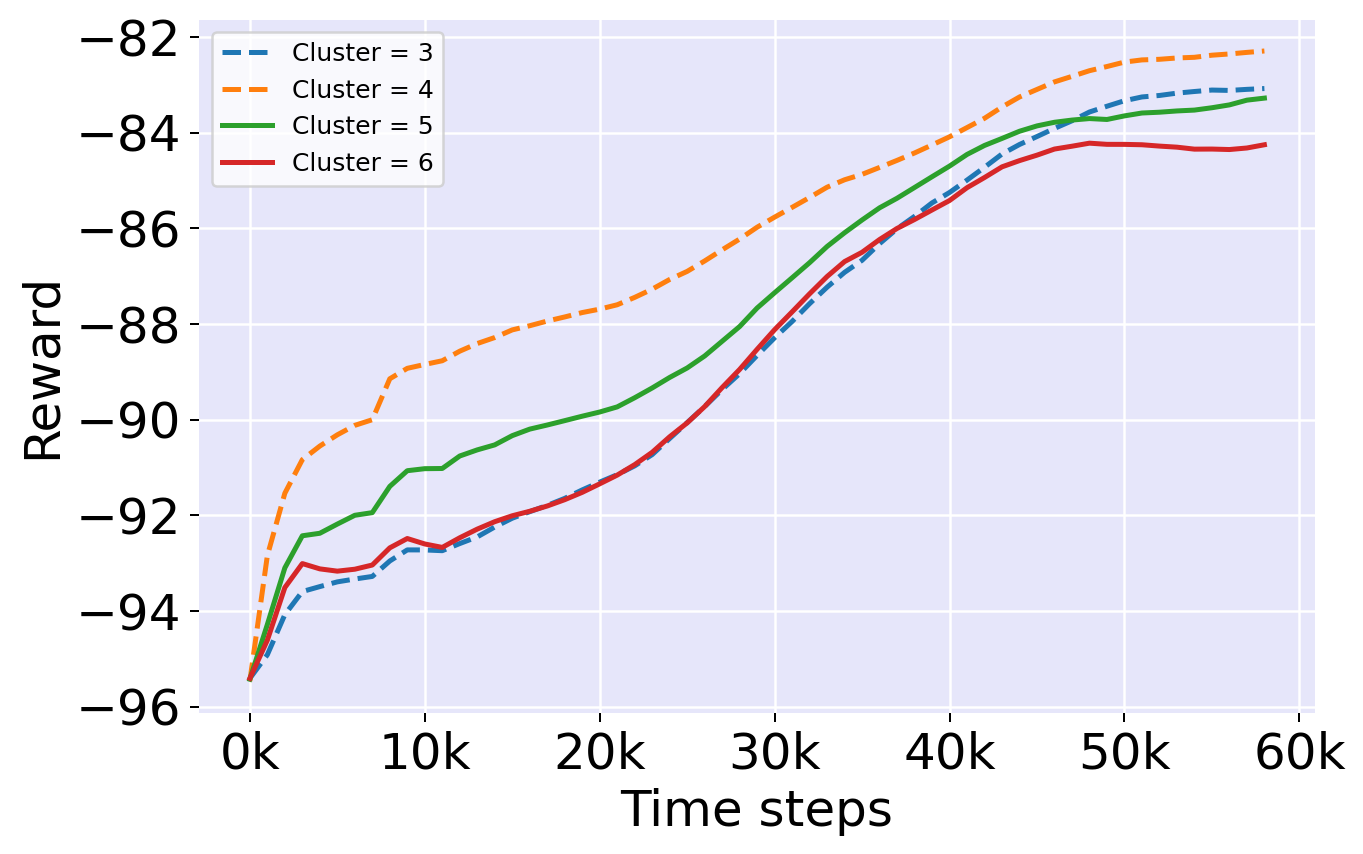}
  \caption{Reward Curve of FedClusterLight with Different Numbers of Clusters $K$}
  \label{fig:Fedclusterlight_curve}
\end{figure}

We conducted a sensitivity analysis on the real-world network and clarified how to select the best $K$ for different networks. Fig.~\ref{fig:Fedclusterlight} shows the cluster distributions produced by FedClusterLight after training with different numbers of clusters. The results indicate that FedClusterLight can capture adjacency relationships between certain intersections. The main groups remain quite stable across different $K$ values, such as intersections at the corners or those on the main road. As $K$ increases, some new groups may emerge, which aligns with our intuition. To select the optimal $K$ for training, we conducted a preliminary experiment, as shown in Fig.~\ref{fig:Fedclusterlight_curve}, which indicates that the algorithm achieves its best performance when $K=4$.

\begin{figure}[H]
        \centering
        \begin{minipage}[c]{0.23\textwidth}
		\centering
		\includegraphics[width=\textwidth]{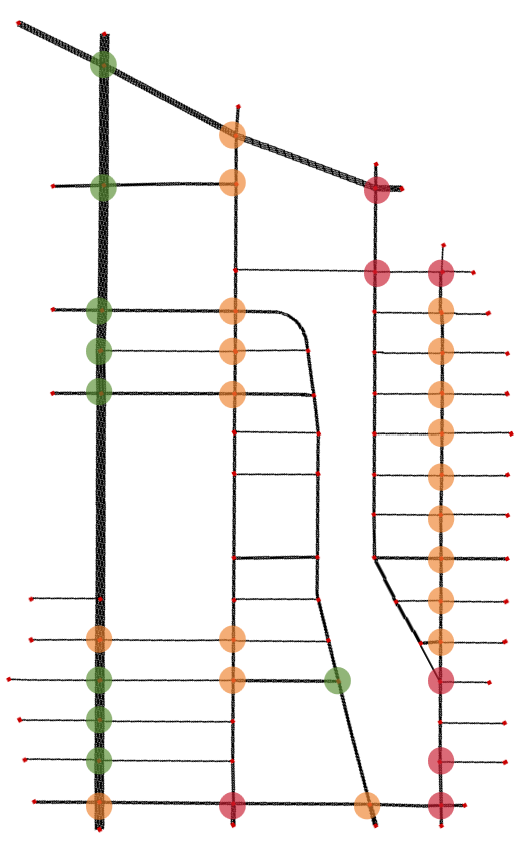}
		\subcaption{Overall cluster distribution.}
	\end{minipage} 
	\begin{minipage}[c]{0.23\textwidth}
		\centering
		\includegraphics[width=\textwidth]{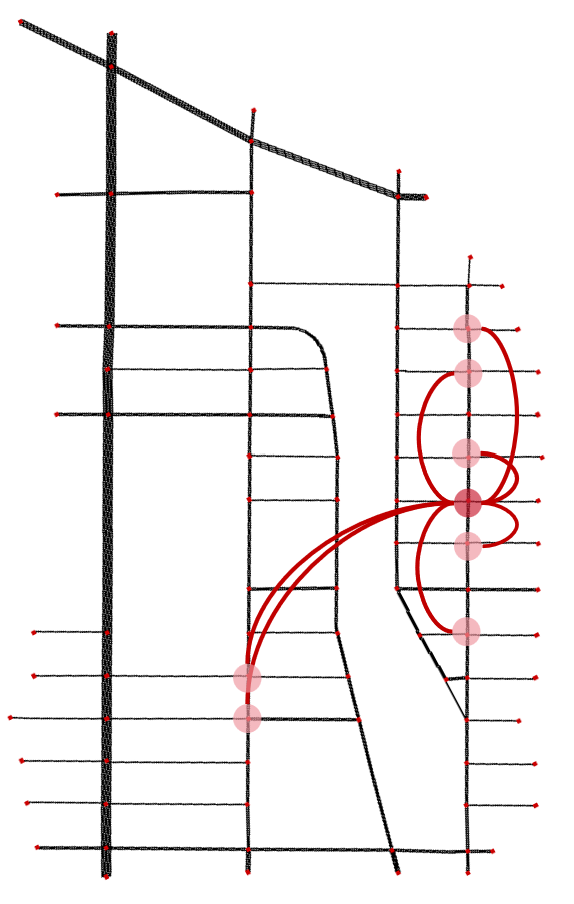}
		\subcaption{Cluster group 1}
            \label{fig:fedfomo-1}
	\end{minipage} 
	\centering
	\begin{minipage}[c]{0.23\textwidth}
		\centering
		\includegraphics[width=\textwidth]{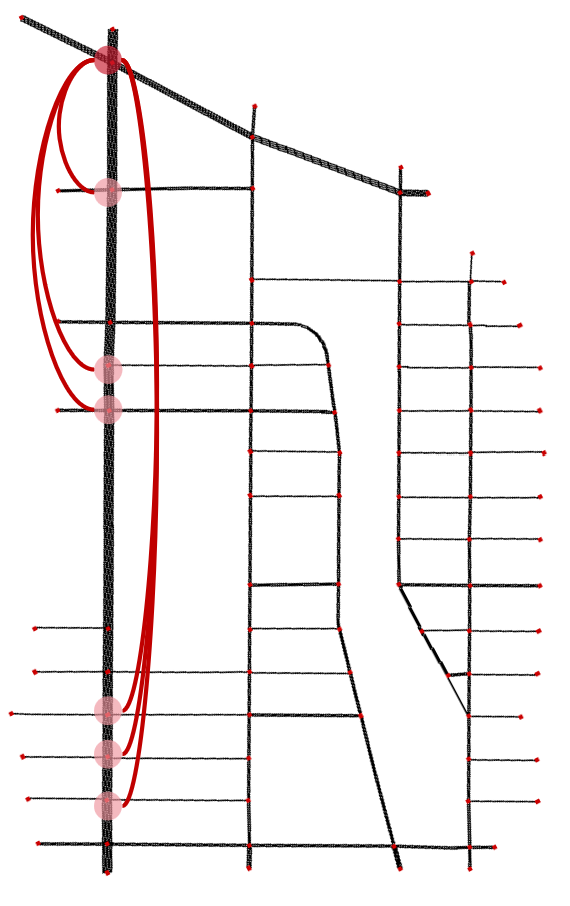}
		\subcaption{Cluster group 2}
            \label{fig:fedfomo-2}
	\end{minipage} 
        \begin{minipage}[c]{0.23\textwidth}
		\centering
		\includegraphics[width=\textwidth]{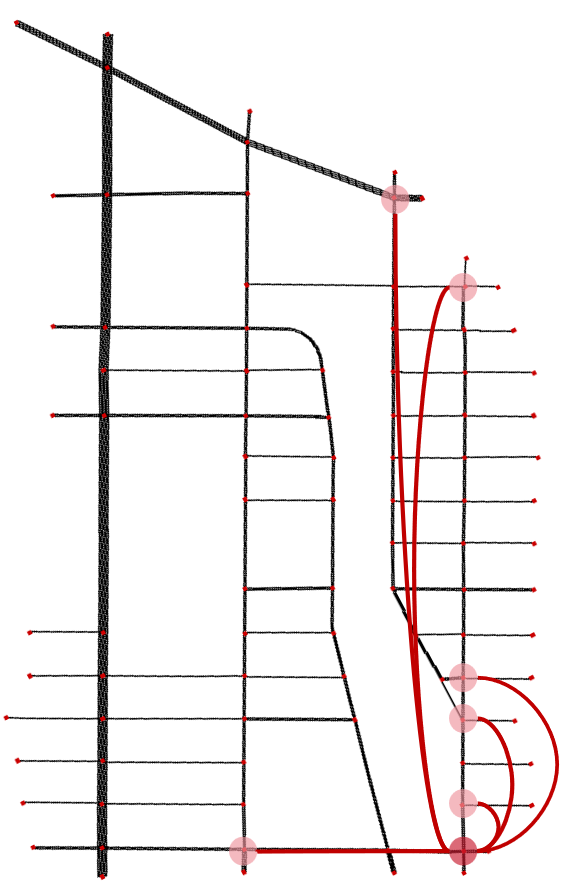}
		\subcaption{Cluster group 3}
            \label{fig:fedfomo-3}
	\end{minipage} 
        \caption{Cluster distribution of FedFomoLight in real-world network}
	\label{fig:cluster_sensity_2}
\end{figure}

Fig. \ref{fig:cluster_sensity_2} presents the hierarchical clustering results of FedFomoLight after training, obtained from a parameter-compression-based matrix. FedFomoLight identifies correlations between model parameters and the traffic network topology, grouping intersection agents accordingly. Cluster group 1 (Fig.~\ref{fig:fedfomo-1}) includes intersections along Amsterdam Avenue and Manhattan Avenue. Cluster Group 2 (Fig.~\ref{fig:fedfomo-2}) consists of intersections along Broadway, while Cluster Group 3 (Fig.~\ref{fig:fedfomo-1}) represents the intersection at the network's corner. Additionally, we randomly selected three intersections and visualized their six most similar counterparts. The results indicate that FedFomoLight captures adjacency relationships among intersections and highlight similarities in their structural features.

\section{CONCLUSION}
\label{conclusion}
This paper introduces an HFRL framework for ATSC, addressing the heterogeneity challenges commonly encountered in large traffic networks using FRL. By integrating hierarchical personalization into FRL, intersections with similar traffic characteristics can share model parameters more effectively, enhancing traffic signal coordination under diverse conditions.

We propose two methods, FedFomoLight and FedClusterLight, to show how personalized aggregation and clustering improve performance over standard federated methods (e.g., FedAvg) and both centralized and decentralized MARL baselines. Experimental evaluations on both synthetic (3×3 and 5×5 grid) and real-world road networks show that these HFRL approaches significantly reduce travel times and waiting times compared to decentralized or globally aggregated baselines, especially in real-world settings. Further analysis reveals that the grouping patterns of FedFomoLight and FedClusterLight strongly correlate with network topology and traffic demand distribution, highlighting their adaptability to heterogeneity.

These results demonstrate the practical benefits of HFRLs: reduced communication overhead, improved scalability for large-scale urban systems, and greater flexibility in handling diverse traffic conditions. Future studies could incorporate the following directions:

\begin{enumerate}[label=(\arabic*)]
    \item Extend the framework to consider additional road users, such as pedestrians at signalized intersections. High pedestrian volumes in urban areas necessitate their inclusion in traffic signal control.
    \item Incorporate additional hierarchical layers (e.g., subregional or corridor-level coordination) into the HFRL framework. This could enhance effectiveness in large-scale traffic networks.
    \item Develop methods to handle sudden disruptions (e.g., road closures, accidents, extreme weather) and enhance model robustness and resilience in atypical scenarios. This helps prevent disruptions at one intersection from negatively impacting the performance of the overall system.
    \item Incorporate fairness-aware learning objectives into the HFRL framework. Future studies could design reward functions or aggregation strategies that ensure intersections receive equally effective models, preventing systematic bias where certain locations experience persistently longer delays.

\end{enumerate}


\section*{Acknowledgments}

This work is sponsored by NSF CPS-2038984 and NSF ERC-2133516.

\bibliographystyle{elsarticle-num-names} 
\bibliography{references}

\end{document}